\pdfoutput=1
\documentclass[12pt]{spieman}  

\usepackage{amsmath,amsfonts,amssymb}
\usepackage{graphicx}
\usepackage{setspace}
\usepackage{tocloft}
\usepackage{epstopdf}

\usepackage{algorithm}
\usepackage{algorithmic}
\usepackage{multirow}
\usepackage{verbatim}
\usepackage{booktabs}
\usepackage{array}
\usepackage{verbatim}
\usepackage{float}
\usepackage{breakurl}
\usepackage{enumerate}

\title{Identifying Designs from Incomplete, Fragmented Cultural Heritage Objects by Curve-Pattern Matching}

\author[a]{Jun Zhou}
\author[a]{Haozhou Yu}
\author[b]{Karen Smith} 
\author[c]{Colin Wilder}
\author[a]{Hongkai Yu}
\author[a]{Song Wang \footnote[1]{Corresponding author. Email: songwang@cec.sc.edu.}}
\affil[a]{University of South Carolina, Department of Computer Science \& Engineering, 315 Main Street, Columbia, SC 29208}
\affil[b]{University of South Carolina, South Carolina Institute of Archaeology and Anthropology, 1321 Pendleton Street, Columbia, SC 29208}
\affil[c]{University of South Carolina, Center for Digital Humanities, 1322 Greene Street, Columbia, SC 29208}

\cftpagenumbersoff{figure}
\cftpagenumbersoff{table} 
\begin{document} 
\maketitle

\begin{abstract}
Study of cultural-heritage objects with embellished realistic and abstract designs made up of connected and intertwined curves crosscuts a number of related disciplines, including archaeology, art history, and heritage management. However, many objects, such as pottery sherds found in the archaeological record, are fragmentary, making the underlying complete designs unknowable at the scale of the sherd fragment. The challenge to reconstruct and study complete designs is stymied because 1) most fragmentary cultural-heritage objects contain only a small portion of the underlying full design, 2) in the case of a stamping application, the same design may be applied multiple times with spatial overlap on one object, and 3) curve patterns detected on an object are usually incomplete and noisy. As a result, traditional curve-pattern matching algorithms, such as Chamfer matching, may perform poorly in identifying the underlying design. In this paper, we develop a new partial-to-global curve matching algorithm to address these challenges and better identify the full design from a fragmented cultural heritage object. Specifically, we develop the algorithm to identify the designs of the carved wooden paddles of the Southeastern Woodlands from unearthed pottery sherds. A set of pottery sherds, curated at Georgia Southern University, are used to test the proposed algorithm, with promising results.

\end{abstract}


\begin{spacing}{1}   

\section{Introduction}\label{sec:intro}

The archaeological record is filled with fragmentary objects of bone, pottery, shell, stone, wood, and cloth variously embellished with realistic and abstract designs. These designs may include figural imagery such as that seen on ancient Maya~\cite{Reents1994} and Greek pottery vessels~\cite{robertson1992art} or the carved marine shell gorgets of late prehistory in North America~\cite{Phillips_shell}. They may also include geometric designs such as those found on Ancestral Pueblo wares~\cite{Huntley_book}. Such imagery also includes maker's marks and seals placed on objects manufactured for markets. Humanities and social science scholars have put these designs to many uses including building chronologies, tracking trade networks, reconstructing aspects of style and the creative process, exploring issues of emulation and resistance, and understanding the creation and expression of identity. 

Without question, most of these topics are best addressed using complete designs rather than design fragments. This is especially the case when reconstructing decorative style is a key part of the research agenda. Such research benefits from the assembly of the largest possible design corpus~\cite{Knight2012}. Traditionally, complete designs are composed using whole artifacts; fragments of designs are then identified as belonging to complete compositions manually by visual assessment\cite{gamble2015archaeology}. The smaller the fragment of design preserved or the more diverse the design corpus, the more difficult it is to match a fragment to a complete composition. The task of matching design fragments to whole designs can be highly time consuming, requiring months or even years of daily effort to identify the fragments of certain complete compositions. As a result, millions of broken cultural heritage objects stored in museums around the world remain unstudied from a design perspective, and large numbers of decorated objects found in the archaeological record contribute little to our understanding of style, production and use, and meaning.

Computer-aided identification of the designs from fragmented cultural objects has attracted great interest among archaeologists and computer scientists in recent years~\cite{halir1999automatic,kampel2007rule}. In this paper, we take pottery sherds found on archaeological sites in the heartland of the paddle-stamping tradition of southeastern North America as our case study, and develop a new computer-vision algorithm to identify the underlying carved wooden paddles impressed on pottery from the Carolinas to the Gulf Coast. 

\begin{figure}[htbp]
\begin{center}
\begin{tabular}{c}
   \includegraphics[width=1\linewidth]{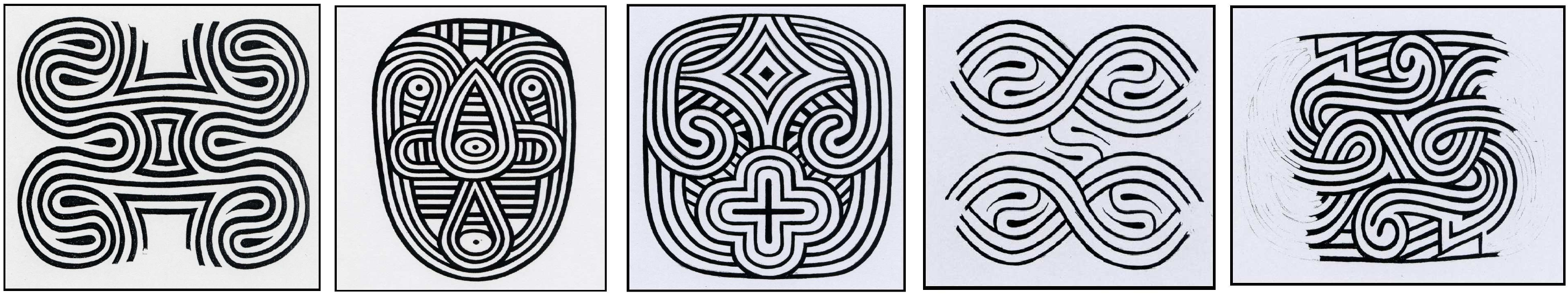}
\end{tabular}
\end{center}
   \caption{Five paddle designs reconstructed by Frankie Snow. Original design reproduced with permission, courtesy of Frankie Snow, South Georgia State College.}
\label{fig:sample designs}
\end{figure}

Elaborately carved wooden paddles of the Southeastern Woodlands, a small fraction of which are shown in Fig.~\ref{fig:sample designs}, represent an ancient Native American art form of the first order, one with rules of stylistic design and technical execution that were taught in communities of practice and passed on from one generation to the next. Every community would have had at least one paddle maker and numerous paddles at their disposal as they gathered to produce pottery vessels. The carved pottery-paddle craft, began in southeastern North America with carved checkered and parallel linework around 500 BC~\cite{stephenson2002aspects} and persisted into the 19th century among some Cherokee potters, making it a craft with deep history in the Southeast~\cite{Holmes1903,Riggs2003}. The ornate, curvilinear paddle impressions on countless pottery sherds of the Swift Creek style tradition made ca. AD 350 to AD 650, at the artistic height of the craft, frame our case study. However, our technical methodology can be applied to carved paddle designs from any subset of the paddle-craft tradition. As demonstrated by Broyles~\cite{Bettye68} and Snow~\cite{snow1975swift}, two archaeologists who spent considerable time reconstructing designs, the research possibilities uniquely presented by paddle design studies are anthropologically significant. For example, our understanding of social and geographical networks, that is, the movement of ideas, people, pots, and paddles across the landscape, is richer as a result of research into the distribution of these unique paddle stamped designs~\cite{wallis2011}.

\begin{figure}[htbp]
	\begin{center}
		\begin{tabular}{c}
			\includegraphics[width=0.7\linewidth]{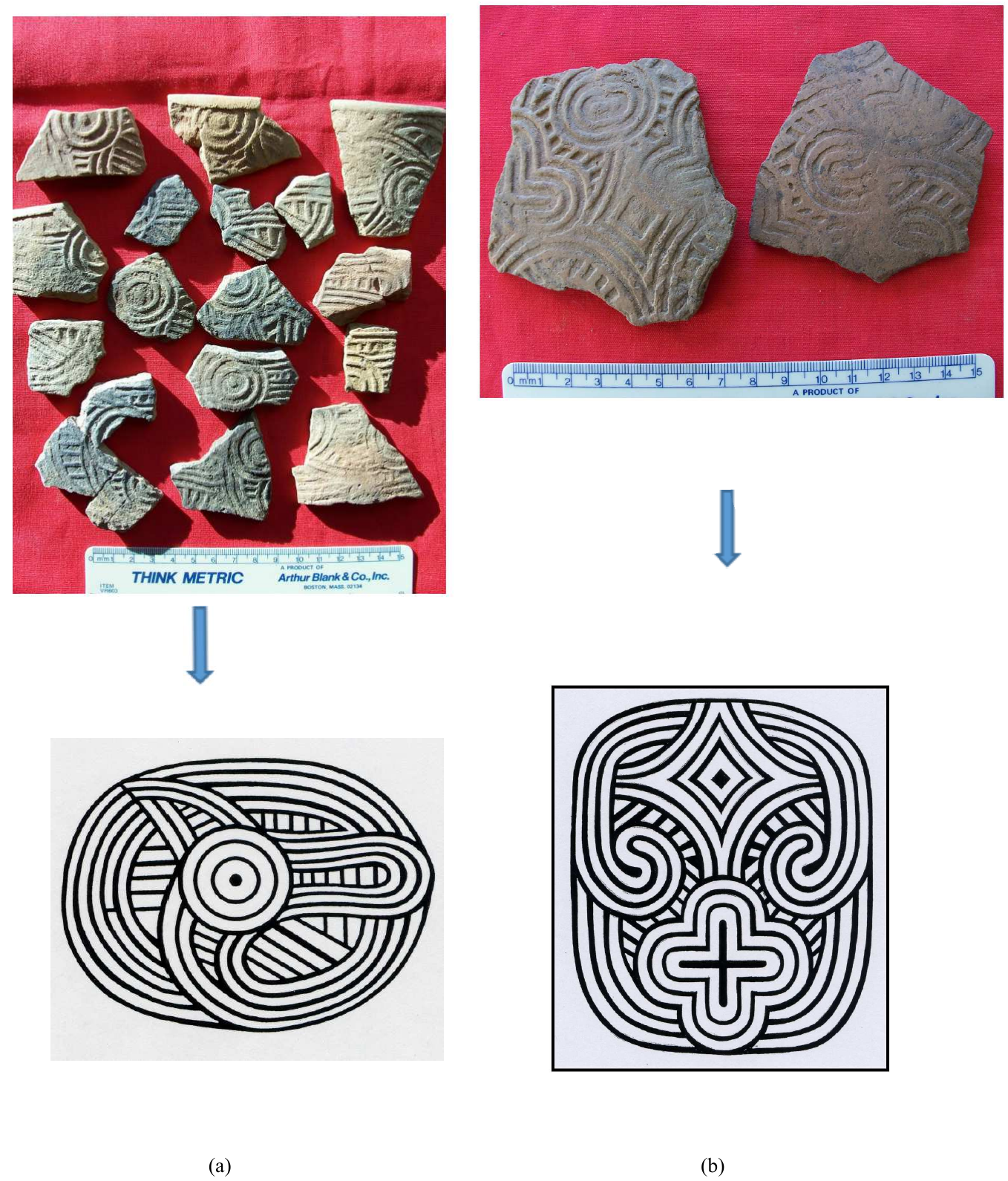}
		\end{tabular}
	\end{center}
	\caption{Sample pottery sherds (top) and their underlying wooden paddle designs (bottom). Two pottery sherds in (b) contain a composite pattern, resulting from the multiple applications of the carved paddle with partial spatial overlaps. Original designs reproduced with permission, courtesy of Frankie Snow, South Georgia State College.}
	\label{fig:sample pottery sherds}
\end{figure}

As shown in Fig.~\ref{fig:sample pottery sherds}, designs carved onto these wooden paddles are primarily composed of connected and intertwined curved lines. The same paddle is usually applied to many different locations on the pottery vessel's exterior surface to achieve the desired decorative effect before the vessel is fired. Also, the same paddle may be applied to many different vessels, fragments of which end up as sherds in the archaeological record. Identifying the full curvilinear paddle design from fragmentary sherds is a highly challenging problem. First, each sherd only contains a small portion of the underlying full paddle design. Second, the available sherds rarely come from the same vessel, and it is difficult to assemble them into large pieces for more complete curve patterns. Third, one carved paddle may be applied multiple times on the pottery surface with spatial overlap, what archaeologists have come to call overstamping. As a result, a sherd may contain a \emph{composite} pattern, i.e., a small fragment of multiple, partially overlapping copies of the same design, as shown in Fig.~\ref{fig:sample pottery sherds}(b). Such a composite pattern is not simply a portion of the full design. Therefore, matching it to the underlying full design is not a simple partial-to-global matching problem~\cite{5995588}. Finally, curve patterns detected on sherds may be incomplete or very noisy due to the gap when applying a planar carved paddle onto a curved pottery surface and to the erosion of sherd surfaces over thousands of years.

In this paper, we develop a new partial-to-global curve matching algorithm for identifying carved paddle designs from pottery sherds by addressing these challenges. More specifically, we extract the curve patterns from a sherd and then match it to each known design in a database and return the best matched designs. The proposed algorithm can automatically recognize whether or not the pattern on a sherd is a composite one and identify multiple components of the composite pattern that correspond to the multiple copies of the same design. In our experiments, we test the proposed algorithm on a set of sherds with a subset of known paddle-stamped designs from the heartland of the paddle-stamping tradition. We achieved a CMC (Cumulative Matching Characteristics) rank-1 rate of 46\% and a CMC rank-2 rate of 65\%, which are much better than several other traditional matching algorithms.

The remainder of this paper is organized as follows. Section~\ref{sec:previous work} reviews the related work. Section~\ref{sec:proposed method} introduces the proposed algorithm for matching the curve pattern on a sherd with known designs. Section~\ref{sec:experiments} introduces the collected test data and the experiment results, followed by a brief conclusion in Section~\ref{sec:conclusion}.

\section{Previous Work}\label{sec:previous work}

Many previous studies on computer-aided processing of archaeological fragments, such as pottery sherds, were focused on classifying whether different fragments come from the same vessel. Classification results are then used to aid the 3D reconstruction of the underlying whole object, such as a full vessel~\cite{cooper2001assembling,gilboa2004towards}.
  Color and texture information have been widely used for fragment classification. Qi and Wang~\cite{li2010kernel} developed texture-based methods for sherd classification by using Gabor wavelet transformation and a non-supervised kernel-based fuzzy clustering algorithm. 
Smith et al.~\cite{smith2010classification} proposed a ceramic sherd classification method based on color and texture features. In particular, it measured color similarity based on the joint probability distribution of the color channels. The method constructed a color histogram in the 3D RGB space, and mesured the texture similarity by using a new texture descriptor similar to geometric total variation energy (TVG) concept proposed by Burchard~\cite{burchard2002total}. Makridis and Daras~\cite{makridis2012automatic} extracted local color and texture features from the front and back views of sherds, and then combined all the local features using the bag-of-words technique. K-Nearest Neighbors (KNN)  algorithm was then used for classification. Rasheed~\cite{rasheed2015archaeological} detected the shared RGB colors and concurrent texture features between different archaeological fragments for classification. 

Many geometric features were also used to classify archaeological fragments. Roman-Rangel~\cite{DBLP:conf/accv/Roman-RangelJA14} developed a potsherd categorization system using the Scale Invariant Feature Transform (SIFT) features and the spin images in 3D space. Bag-of-words technique was then used to combine all the local features, followed by principal component analysis (PCA) to further reduce the feature dimensions. Maiza and Gaildrat~\cite{karasik20083d} proposed an algorithm to use the 3D surface geometry for fragment classification. Karasik~\cite{karasik2011computerized} assumed that the pottery sherds were from spherically symmetric vessels and used this shape prior for fragment classification and 3D reconstruction. Other than fragment classification, many previous works~\cite{cooper2001assembling, cohen2010virtual} were focused on developing algorithms to assemble sherds into larger pottery pieces, or the whole vessel, by fitting the boundary shape of sherds. 

While the proposed work can also be treated as a sherd classification problem by classifying sherds according to different designs, it deviates significantly from the works described above. In this work, sherds with the same design are rarely from the same vessel, meaning often it is impossible to reconstruct an entire vessel, or even larger vessel pieces of a vessel. In another words, sherds with the same design are usually from different vessels, with different shapes, sizes, colors and textures. As a result, we could not use the color, texture, and geometric information in this work as in previous fragment classification studies.

From the algorithm perspective, this paper aims to find a match between a partial curve pattern (on a sherd) and a full curve pattern (a design). Partial matching is a long studied problem in computer vision for different applications.  Huttenlocher et al.~\cite{huttenlocher1993comparing} suggested the use of Hausdorff Distance for such pattern matching.  Belongie et al.~\cite{belongie2002shape} proposed a shape context approach for curve-pattern matching by building a log-polar histogram around each sampled curve point and then using this histogram as the feature to match the curve points between two curve patterns. Roman-Rangel~\cite{roman2011analyzing} extended the shape context algorithm to a Histogram of Orientation Shape Context (HOOSC) algorithm by incorporating the orientation measure into the log-polar histograms. Promising results have been reported using HOOSC to analyze the Ancient Maya Glyph collections. Both shape context and HOOSC are invariant to the scaling and rotation between two matched patterns. Chamfer matching algorithm~\cite{liu2010fast,munsell2009fast} is widely used for partial matching of curve patterns. It computes a distance map from the full curve pattern and then slides the partial pattern over the distance map to find the optimal matching location and matching cost. The pre-computing of the distance map can substantially increase the computational efficiency. Brunelli~\cite{Brunelli2009} introduced an image matching method that uses a linear spatial filtering algorithm between two patterns by treating one as a convolution mask over the other. However, none of these existing methods considered the composite patterns that are common in this work, where the partial pattern is a fragment of multiple, partially overlapping copies of the same design. In this case, the partial pattern is actually not simply a portion of the full pattern, which is an important assumption in most existing partial matching methods. In Section~\ref{sec:experiments}, we include Chamfer matching, image matching, shape context and HOOSC as the comparison methods in our experiments and evaluate their performances.

Also related to our proposed work is Opelt et al.~\cite{opelt2006boundary}, where a boundary fragment model  was developed for object detection. The basic idea is to identify the discriminative shape features underlying an object boundary through machine learning, and then to determine whether a curve fragment extracted from an unseen object belongs to the desired object boundary using the learned features. This method, however, cannot be used to address our problem of identifying designs from sherds. 1) The curves in the designs, as shown in Figs.~\ref{fig:sample designs} and~\ref{fig:sample pottery sherds}, are usually of very simple shape, making it difficult to discern discriminative shape features from curve fragments. 2) Only a small portion of design is present on a sherd, which may not be sufficient to make a matching to the underlying design using the boundary fragment model. In Opelt et al.~\cite{opelt2006boundary}, the boundary-fragment model was developed for object detection, where it is assumed that most of the object boundary is available in the form of disjoint curve fragments. 3) The boundary-fragment model does not consider the composite pattern that is common in our work. A composite pattern with many curve intersections on a sherd will substantially increase the difficulty of extracting informative boundary fragments.

\section{Proposed Method}\label{sec:proposed method}

\begin{figure}[htbp]
\begin{center}
   \includegraphics[width=1\linewidth]{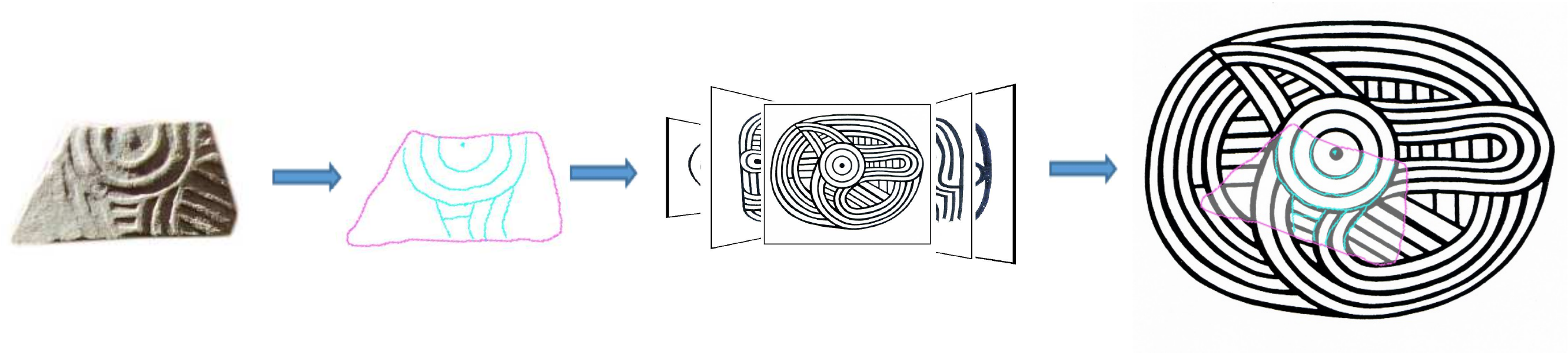}
\end{center}
   \caption{An illustration of the procedure of identifying the underlying design for a sherd: first extracting the curve pattern on the sherd, which is then matched to each design in a database of known designs for identifying the best matched design. Original design reproduced with permission, courtesy of Frankie Snow, South Georgia State College. }
\label{fig:non-composite sherd-to-design matching procedure}
\end{figure}

As in previous matching algorithms, the key step is to quantitatively define a matching distance or matching score between a sherd and each design drawn from the database of known designs. Only a small number of designs with lowest matching distances or highest matching scores can then be considered as the design used in the sherd and they are presented to archaeologists for a final decision. Figure~\ref{fig:non-composite sherd-to-design matching procedure} shows a sample sherd, its curve pattern, a design drawn from the design database and the sherd-to-design matching result.  

 Typically, pottery sherd images are color images that are taken by archaeologists or curators using a camera held nearly perpendicular to these sherds. A ruler was placed by these sherds to indicate their actual size, as shown in Fig.~\ref{fig:sample pottery sherds}. Paddle designs are manually constructed by highly knowledgeable and experienced archaeologists after finding a number of sherds with the same design. The curves in the produced design images, including the curve geometry and width, reflect the ones carved on the original paddle and displayed on the original pottery. Two examples of the design images are shown in Fig.~\ref{fig:sample pottery sherds}. All the design images are also provided with their actual sizes, e.g., in centimeters. For the sherds and designs studied in this paper, size is a discriminative feature -- even if two designs look exactly the same except for their size, they are still different designs because they correspond to two paddles of different sizes.  In another words, the proposed matching between sherd and design is not scale invariant, and we resize all the sherd and design images to have a uniform DPI (dots per inch) before curve extraction and matching.

\begin{figure}[htbp]
\begin{center}
   \includegraphics[width=1\linewidth]{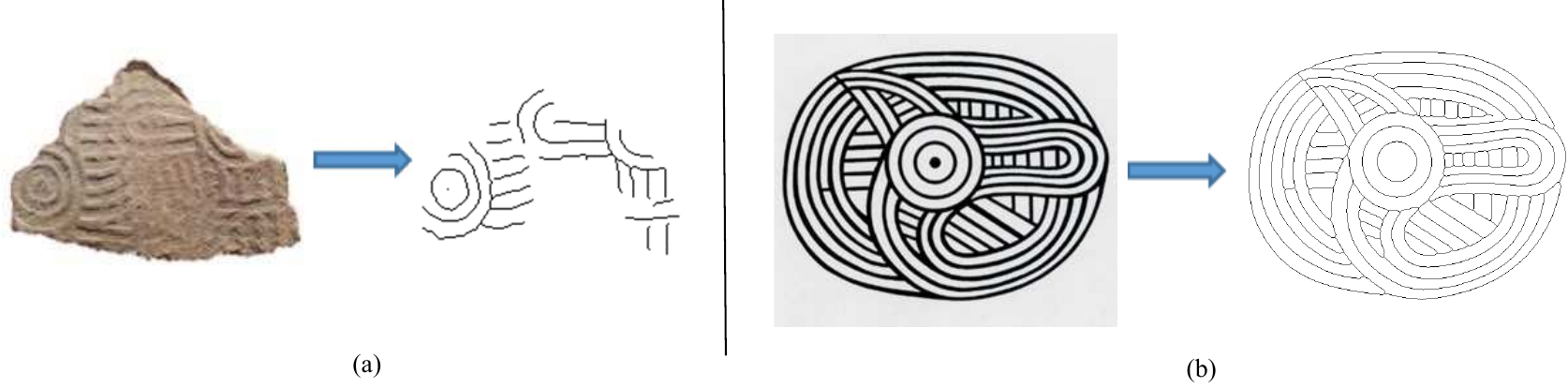}
\end{center}
   \caption{An illustration of curve extraction from a sherd and a design. (a) Curve extraction from a sherd. (b) Curve extraction from a design. Original design reproduced with permission, courtesy of Frankie Snow, South Georgia State College.}
\label{fig:curve from a design}
\end{figure}

While ideally the curve width can be used as an important clue in matching the sherd and a design, we try not to use the curve-width information in this paper because it is very difficult to accurately measure the curve width from a deteriorated sherd surface. Therefore, in this paper, we first extract one-pixel wide curves from both the sherd images and the binary paddle stamp design images, as shown in Fig.~\ref{fig:curve from a design}, and the matching distance is then defined based only on the one-pixel wide curves. Chamfer matching is one of the most widely used and effective algorithms used for partial-to-global curve pattern matching~\cite{barrow1977parametric} . However, the classical Chamfer matching requires one pattern to be a portion of the other, which is not true in this paper when the curve pattern on the sherd is a composite one. To address this issue, we propose to develop a new algorithm that can automatically identify multiple components of the composite pattern extracted from the sherd. In the following, we first discuss the curve pattern extraction from the sherd and design images. Then we briefly review the classical Chamfer matching algorithm. Finally we introduce the proposed new partial-to-global matching algorithm.

\subsection{Curve Extraction}\label{Curve Extraction}

\begin{figure}[htbp]
	\begin{center}
		\includegraphics[width=1\linewidth]{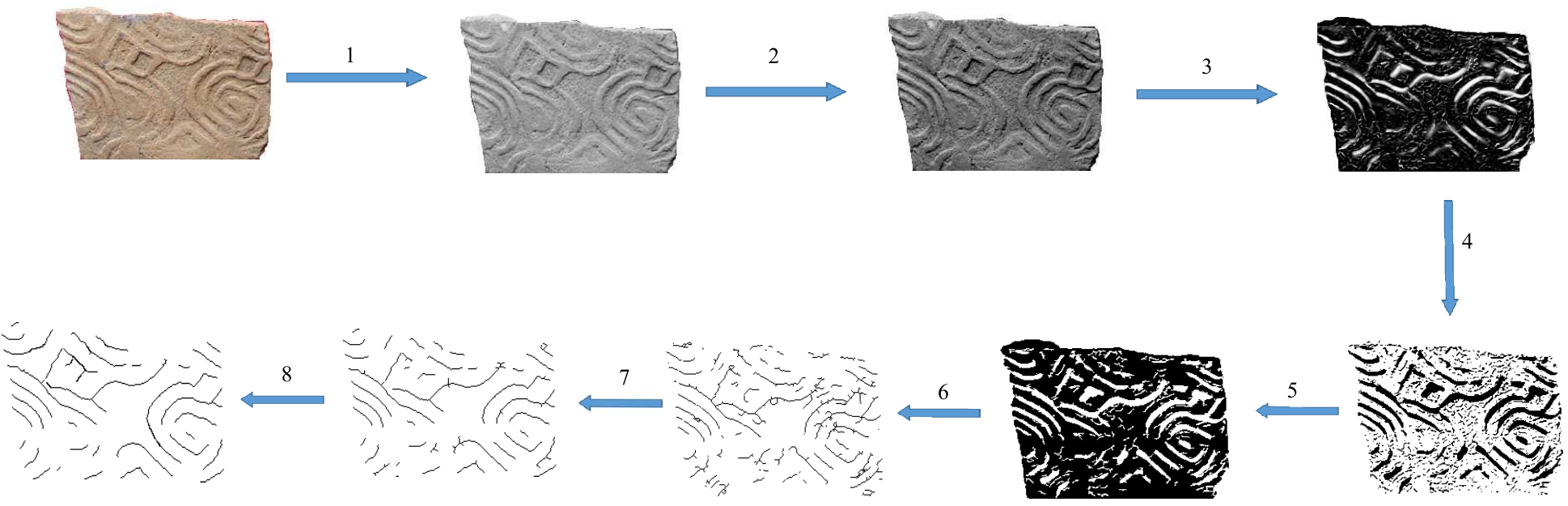}
	\end{center}
	\caption{An illustration of curve pattern extraction from a sherd. 1) Converting color image to gray-scale image. 2) Image enhancement. 3) Ridge detection. 4) Thresholding for binary ridge image. 5) Noise removal.
			6) Thinning. 7) Short branch removal. 8) Manual refinement (if needed).}
	\label{fig:curve extraction}
\end{figure}

In this work, sherd images are taken on a background with a uniform color that is not present in the sherd (red in Fig.~\ref{fig:sample pottery sherds}). This way, we can easily remove the background and focus on the foreground region of the sherd. We take the following steps to extract the one-pixel-wide curve pattern from the foreground region, as illustrated in Fig.~\ref{fig:curve extraction}. 
\begin{enumerate}[{1)}]
	\item Convert the color sherd image to a gray-scale image using the standard MATLAB function ${\tt rgb2gray}$  and its default parameters.
	\item Enhance the gray-scale image by increasing the contrast between the stamped curves and the nearby intact sherd surface. We use MATLAB function ${\tt imadjust}$ and its default parameters for this step. 
	\item Apply a ridge detector based on a multiscale Hessian filter (HBF)~\cite{frangi1998multiscale}. We used the MATLAB/C/C++ based implementation by Dirk-Jan Kroon at the University of Twente for this step\footnote[1]{https://www.mathworks.com/matlabcentral/fileexchange/24409-hessian-based-frangi-vesselness-filter}. 
	The HBF is calculated on 10 exponentially distributed scales with an inter-scale ratio of 2. For the other three parameters in this implementation, we set $\alpha=0.5$, $\beta=0.5$ and, $\gamma=15$.  $\alpha$, $\beta$ and $\gamma$ are thresholds which control the sensitivity of the filter to measure blob, plate- and line-like structures in the application. 
	\item Detect binary ridges by thresholding the image resulting from Step 3). We use the MATLAB function ${\tt im2bw}$ with a threshold $0.2$ for this step. 
	\item Remove isolated noise dots (with area less than 10 pixels) from the binary ridge image, using MATLAB function ${\tt bwareaopen}$ with a threshold $10$ for this step. 
	 \item Perform a thinning operation using MATLAB function ${\tt bwmorph}$ to get one-pixel-wide curves. 
	 \item Remove small branches (less than 10 pixels) by using the MATLAB function ${\tt findendsjunctions}$ included in LineSegments package\footnote[7]{http://www.peterkovesi.com/matlabfns/}. 
	 \item Manually refine the curves by adding long missing curves and removing long false-positive curves on the images with very poor curve-detection results after Step 7).
\end{enumerate}	

\begin{figure}[htbp]
\begin{center}
\includegraphics[width=1\linewidth]{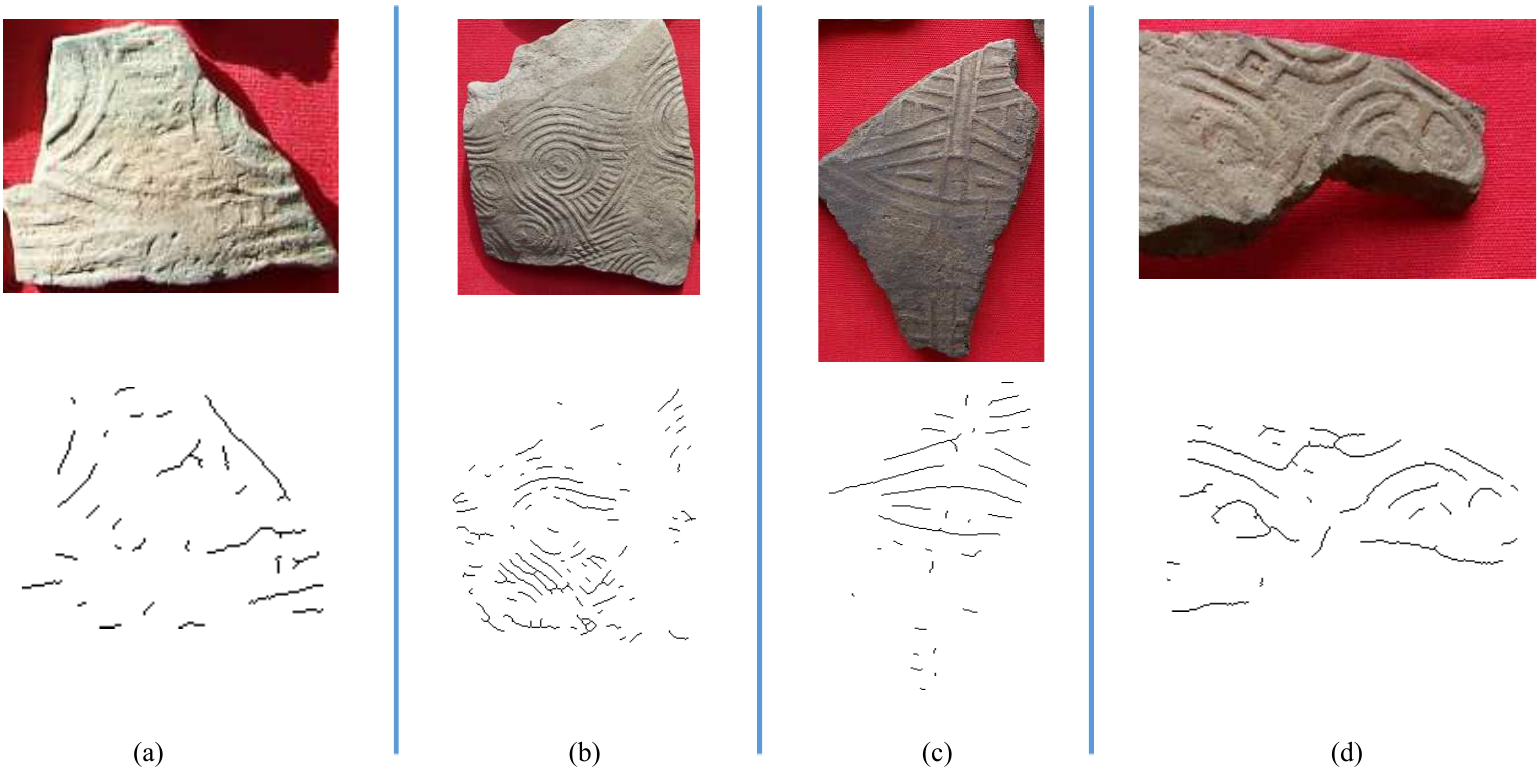}
\end{center}
   \caption{(a–d) Curve extraction results (bottom) on four samples sherds (top) using automatic image processing, i.e., Steps 1) through 7). They need substantial manual refinement.}
\label{fig:failure cases on curve extraction}
\end{figure}

The sherd images currently used in this work were taken by hand-held cameras. Sherd surface deterioration, curved surfaces, camera perspectives, image deformation, and improper lighting make it very difficult to extract 
the curve pattern on a sherd  using a fully automatic algorithm. Poor curve extraction results on several sample sherd images, using only Steps 1) through 7), are shown in Fig.~\ref{fig:failure cases on curve extraction}. These results need substantial manual refinement in Step 8) before they can be used for matching and design identification. 
For the sherd images tested in our experiments (Section~\ref{sec:experiments}), about $50\%$ of them need substantial manual refinement in curve extraction. In the future, we expect that a specifically designed calibrated and unified imaging system can collect higher-quality sherd images, from which we can extract high-quality curves without any manual refinement.

The design images are manually constructed gray-scle images where curve patterns have an intensity value and the background has another intensity value.  We use a standard edge-thinning algorithm~\cite{lam1992thinning} to reduce the curve width to one pixel as illustrated in Fig.~\ref{fig:curve from a design}(b). The curve patterns derived from both the sherd images and the design images are one-pixel wide. We call the pixels on the curves \emph{edge pixels} in the remainder of the paper.

\subsection{Chamfer Matching of Curve Patterns}\label{Chamfer Matching algorithm}

Chamfer matching~\cite{barrow1977parametric} is an efficient algorithm that has been successfully used for partial-to-global matching of one-pixel-wide curve patterns. In this paper, we will develop the proposed curve pattern matching algorithm by extending several concepts in Chamfer matching. In this section, we briefly review the traditional Chamfer matching algorithm.

Given two (one-pixel wide) curve patterns $U$ and $V$, the goal of Chamfer matching is to decide whether $U$ is a portion of $V$ and if so, find the transform that matches the partial pattern $U$ to the full pattern $V$. For clarity, in the paper we use $U$ and $V$ to denote the set of 2D coordinates of the edge pixels in these two patterns, respectively. 
The transform $\bf T$ that matches $U$ to $V$ usually consists of a translation ${\bf t}=(t_x, t_y)$, a rotation of angle $\theta$, and a scaling with factor $s$. Chamfer matching is based on the Chamfer distance. Let $U_{\bf T}$ be the partial pattern $U$ after the transform $\bf T$. Aligning $U_{\bf T}$ and $V$, we can define the Chamfer distance between them as
\begin{equation}
\label{eq:chamfer distance}
d_{CM}(U_{\bf T},V) = \frac{1}{|U_{\bf T}|}\sum_{{\bf u} \in U_{\bf T}}\min_{{\bf v}\in V}\left \| {\bf u} - {\bf v} \right \|_2, 
\end{equation} 
where ${\bf u} \in U_{\bf T}$ indicates all the edge-pixel coordinates ${\bf u} $ in the transformed partial pattern $U_{\bf T}$ and ${\bf v}\in V$ indicates all the edge-pixel coordinates ${\bf v}$ in the curve pattern $V$. $|U|$ is the total number of edge pixels in the partial pattern $U$. Eq.~(\ref{eq:chamfer distance}) actually finds the nearest edge-pixel coordinate in $V$ for each edge-pixel coordinate in $U_{\bf T}$, records its Euclidean distance $ \left \| {\bf u} - {\bf v} \right \|_2$ and finally averages over all the edge-pixel coordinates in $U_{\bf T}$. 

By trying all possible transforms $\bf T$'s, the transform ${\bf T}^*$ for the best matching can be determined by 
\begin{equation}
\label{eq:best_location}
{\bf T}^*=\arg\min_{\bf T} d_{CM}(U_{\bf T},V).
\end{equation} 
The optimal transform ${\bf T}^*$ leads to the Chamfer matching between $U$ and $V$ with matching distance
\begin{equation}
\label{eq:best_distance}
 d_{CM}(U_{{\bf T}^*},V).
\end{equation} 
In practice, we can examine the matching distance  $d_{CM}(U_{{\bf T}^*},V)$ -- if it is larger than a given threshold, we may consider that $U$ is not a partial pattern of $V$; otherwise, we can consider that  $U$ is a partial pattern of $V$ and ${\bf T}^*$ provides the location, orientation and scaling that match $U$ to $V$. In Chamfer matching, we need to search over all possible transform parameters of $\bf T$. Therefore, the reduction of the degrees of freedom in $\bf T$ can substantially reduce the search space and speed up the algorithm. In this paper, we match sherd pattern $U$ to the full paddle design $V$. The matching is not scale invariant and all the sherd and design images have been preprocessed to have a uniform DPI (dots per inch), as discussed at the beginning of this section. Therefore, the transform $\bf T$ in this paper only consists of a translation $\bf t$ and a rotation with angle $\theta$, i.e.,
\begin{equation}
\label{eq:chamfer warping function}
{\bf T}(\bf u) =\begin{pmatrix}
\cos\theta&-\sin\theta\\
\sin\theta&\cos\theta
\end{pmatrix}{\bf u}+\begin{pmatrix}
t_x\\
t_y
\end{pmatrix},
\end{equation} 
where the search range for $\theta$ is $[0^{\circ}, 360^{\circ})$ and the search range for $t_x$ and $t_y$ can be constrained by the size (width and height) of the bounding boxes that tightly cover the sherd pattern $U$ and design $V$.

\begin{figure}[htbp]
\begin{center}
   \includegraphics[width=1\linewidth]{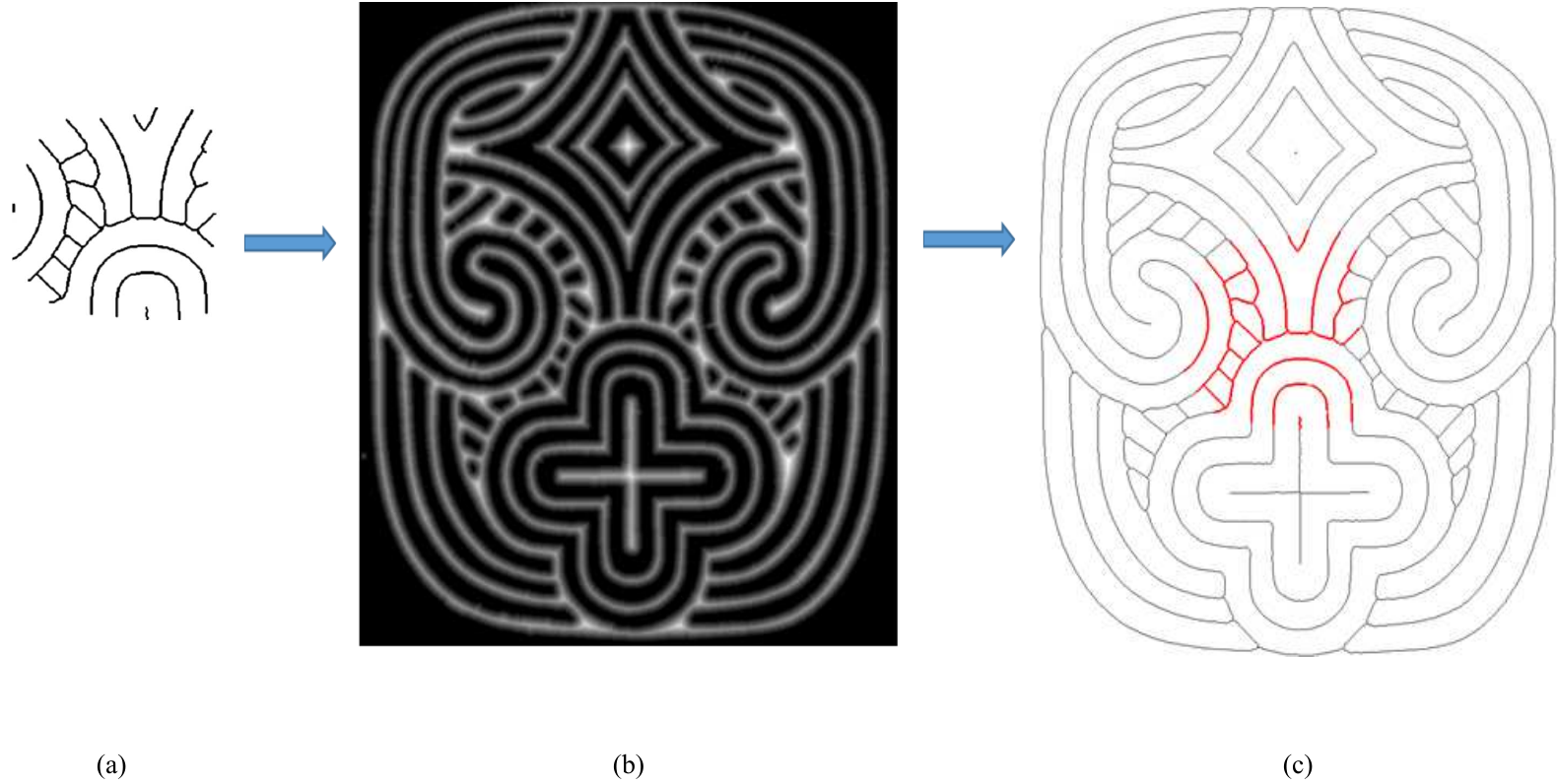}
\end{center}
   \caption{An illustration of the distance map and Chamfer matching. (a) Curve pattern on a sherd. (b) Distance map of a design - brighter pixels indicate higher values in the distance map. (c) Chamfer matching result (in red). From original design by Frankie Snow, South Georgia State College. 
}

\label{fig:distance map}
\end{figure}

Based on Eq.~(\ref{eq:best_location}), we need to calculate the Chamfer distance $d_{CM}(U_{\bf T},V)$ for each possible choice of parameters in transform $\bf T$. In Chamfer matching, this can be accelerated by pre-computing the distance map for $V$ -- the distance map value $M({\bf u})$ at any 2D coordinate ${\bf u}$ indicates the Euclidean distance from $\bf u$ to the nearest coordinate of an edge pixel in $V$. 
For ${\bf v}\in V$, we have $M({\bf v})=0$. This way, Eq.~(\ref{eq:chamfer distance}) can be simplified as
\begin{equation}
\label{eq:chamfer distance-map}
d_{CM}(U_{\bf T},V) = \frac{1}{|U_{\bf T}|}\sum_{{\bf u} \in U_{\bf T}}M({\bf u}) .
\end{equation}  
An example of the distance map built for a design $V$ and its use for computing Chamfer distance is shown in Fig.~\ref{fig:distance map}.

\subsection{Composite Pattern Matching}\label{sec:composite pattern matching}

\begin{figure}[htbp]
\begin{center}
   \includegraphics[width=1\linewidth]{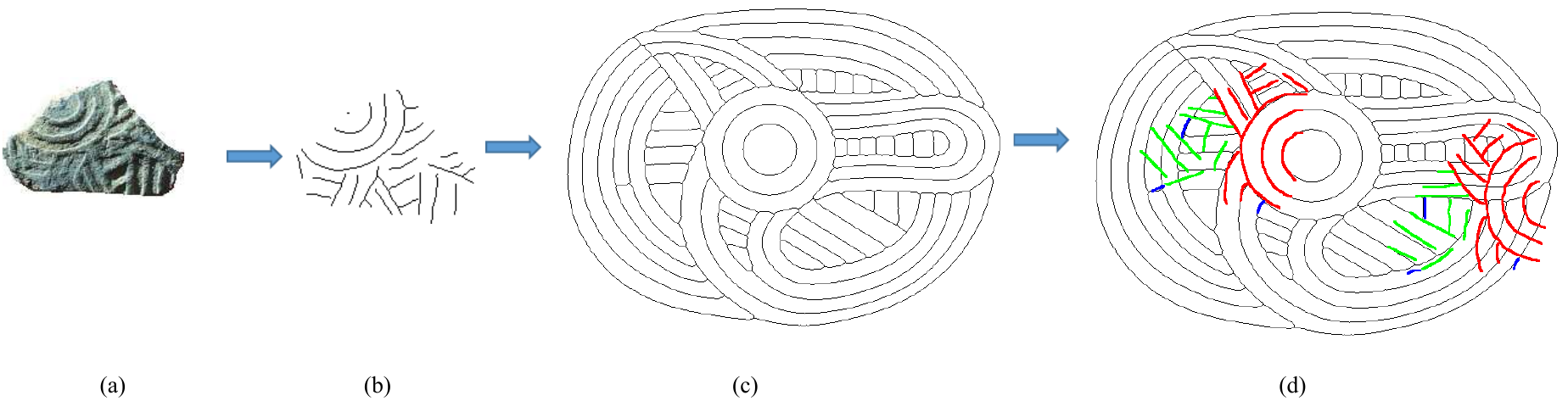}
\end{center}
\caption{An Illustration of a composite pattern, which consists of two components. (a) A sherd with a composite pattern. (b) The extracted composite pattern. (c) The underlying design. (d) Two components (red and green) of the composite pattern matched to different parts of the design, with blue curve fragments shared by two components. From original design by Frankie Snow, South Georgia State College.}

\label{fig:composite pattern curves matching problem}
\end{figure}

The classical Chamfer matching discussed above requires that the pattern $U$ is a portion of the full design $V$, under a transform $\bf T$. In the proposed sherd-to-design matching problem, it basically requires that the sherd partially contains a single copy of the full design. However, in the case of paddle-stamped pottery, the curve pattern on a sherd may be a composite one -- the same carved paddle was applied to the pottery surface multiple times with spatial overlap and the sherd may partially contain multiple, spatially overlapping copies of the same design. An example is illustrated in 
Fig.~\ref{fig:composite pattern curves matching problem}, where a sherd curve pattern consists of two components, corresponding to the two overlapping copies of the same design.
Two components (red and green) of the composite pattern are matched to different parts of the design, with blue curve fragments shared by two components. In this case, the direct application of the traditional Chamfer matching could not find the correct partial-to-global matching between the curve pattern $U$ on the sherd and the design $V$.

To address this problem, we need to allow different parts of the sherd pattern $U$ to be matched to the different parts of the design $V$. Ideally, $U$ can be matched to $V$ by decomposing $U$ into $\{U_1, U_2, \cdots, U_K\}$ such that 
\begin{eqnarray}
	&U=\bigcup_{k=1}^K U_k \label{eq:condition-1} \\
	& U_i\bigcap U_j=\emptyset, \forall i\neq j; i,j =1,2,\cdots, K, \label{eq:condition-2}
\end{eqnarray}
and then each component $U_k$ can be matched as a portion of $V$ with its own transform. This way, we can define the matching distance (or score) between $U$ and $V$ by combining the matching distance (or score)
between each component $U_k$ and $V$.

The above first condition in Eq.~(\ref{eq:condition-1}) reflects the \emph{completeness} of the decomposed pattern components $U_k,k=1,2,\cdots, K$. Considering the possible noise in the sherd pattern,  we simply seek the decomposition that maximizes the completeness 
\begin{equation}
\phi_c(U_1, U_2, \cdots, U_K)=\frac{|\bigcup_{k=1}^K U_k|}{|U|}.
\label{eq:completeness}
\end{equation}

The above second condition in Eq.~(\ref{eq:condition-2}) reflects the \emph{disjointness} of the decomposition. Considering the possibility of shared curve fragments across multiple components, as illustrated in Fig.~\ref{fig:composite pattern curves matching problem},  in this paper we relax this condition to 
\begin{equation}
\phi_d(U_i, U_j)=\frac{|U_i\cap U_j|}{|U_i\cup U_j|}<\eta, \forall i\neq j; i,j =1, 2,\cdots, K,
\label{eq:disjointness}
\end{equation}
where $\phi_d(U_i, U_j)$ is the \emph{disjointness} between two components $U_i$ and $U_j$, and $\eta$ is a preset threshold for the disjointness.

In practice, we also need to limit the number of components $K$. If we over-decompose $U$ to too many very simple curve patterns, e.g., each $U_k$ only contains one edge pixel, then we can always perfectly match each $U_k$ to $V$ with zero Chamfer distance with a translation. Considering that the sherds are highly fragmented pieces of the pottery and it is not common to see a sherd that partially contains more than two copies of the full  design $V$, we only consider the cases of $K\leq 2$ in all our experiments. 

Based on these considerations, the main problem we need to address is to find the optimal decomposition $U_k, k=1,2,\cdots, K$ for $U$ to match the design $V$ and quantify the matching distance or score.  Clearly the possible choices of decomposition are very large given the large number of edge pixels in $U$ and we could not try every possible decomposition to search for the global optimum. In this paper, we use the Chamfer distance between each component of $U$ and $V$ to reduce the search space of decomposition. 

More specifically, for each possible transform ${\bf T}$ consisting of a translation $\bf t$ and a rotation $\theta$, we align $U_{\bf T}$ and the design $V$. Using the distance map $M(\cdot)$ for $V$, we can construct a candidate component $U({\bf T})=U({\bf t}, \theta)\subseteq U$  by collecting all the edge-pixel coordinates $\{{\bf u}|{\bf u}\in U, M({\bf T}({\bf u}))<\alpha\}$, where $\alpha$ is a threshold to determine whether ${\bf u}$ has a corresponding matching edge pixel in $V$ under the transform $\bf T$ and let $d_{CM}(U_{\bf T}({\bf T}), V)$ be the Chamfer distance between this candidate component, after the transform $\bf T$, and $V$. For each translation $\bf t$, we first try all possible values of $\theta$ in  $[0^{\circ}, 360^{\circ})$ and keep the one that leads to the minimum Chamfer distance, i.e.,
\begin{equation}
\theta_{\bf t}^*=\arg\min_{\theta} d_{CM}(U_{\bf T}({\bf t}, \theta), V).
\end{equation}
We construct a candidate component $U({\bf t})=U({\bf t}, \theta_{\bf t}^*)$ for each translation offset $\bf t$ and the matching distance between this candidate component and $V$ is defined as
\begin{equation}
\tilde{d}({\bf t})=d_{CM}(U_{({\bf t}, \theta_{\bf t}^*)}({\bf t}, \theta_{\bf t}^*), V).
\label{eq:candidate component construction}
\end{equation}

After constructing a candidate component at each possible translation $\bf t$, we obtain a new distance map $\tilde{d}$ of the same size as the distance map $M(\cdot)$ for $V$.
We also construct a large number of candidate components from $U$, one for each translation offset $\bf t$. Trying all possible combinations of these candidate components is computationally expensive.  In practice, one can expect that the  candidate components constructed at two neighboring $\bf t$'s are very similar. Therefore, we use a minimum-suppression strategy to further reduce the number of candidate components. Specifically, we find the regional local minimum on
the new distance map $\tilde{d}$ and only keep the candidate components at $\bf t$'s corresponding to these local minimums. 
Assume that the local minimums are found at ${\bf t}_1, {\bf t}_2\cdots, {\bf t}_p$ and their corresponding candidate components are $U({\bf t}_1), U({\bf t}_2),\cdots, U({\bf t}_p)$ respectively. We can consider the $K$ combination of them for final components. As discussed above, we set the actual number of components in $U$ to be $K\leq 2$. Therefore, we limit the search space for the decomposition of $U$ to the following $p+\frac{p(p-1)}{2}$ cases:
\begin{enumerate}
	\item The $p$ cases where $U$ only contains one single component, i.e., $U({\bf t}_1), U({\bf t}_2),\cdots, U({\bf t}_p)$.
	\item The $\frac{p(p-1)}{2}$ cases where $U$ contains two components, i.e., $U({\bf t}_i)$ and $U({\bf t}_j)$, with $i<j$, $i,j=1,2,\cdots, p$.
\end{enumerate} 

\begin{figure}[hp]
\begin{center}
   
   \includegraphics[width=1\linewidth]{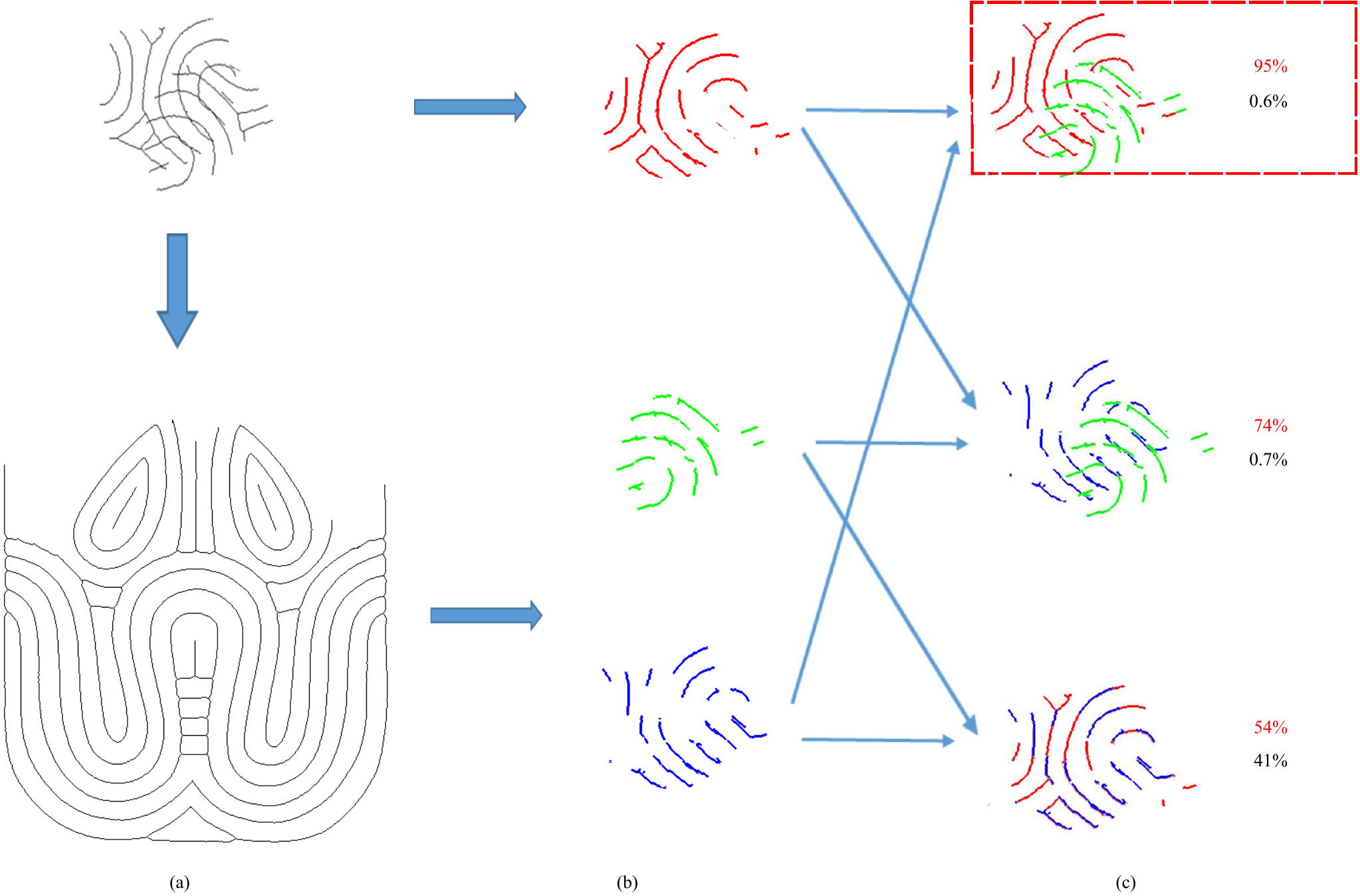}
\end{center}
\caption{The process of combining candidate components for matching to a design ($K=2$). The optimal result is indicated in the red box. (a) Matching a sherd pattern (top) to a design pattern (bottom). (b) Candidate components. (c) Combining candidate components (completeness scores $\phi_c$ shown in red and disjointness scores $\phi_d$ shown in black). From original design by Frankie Snow, South Georgia State College.}
\label{fig:candidate combination}
\end{figure}

This can be extended to the cases where $U$ is decomposed into more than two components, but the size of the search space will substantially increase.
For each of the $p+\frac{p(p-1)}{2}$ cases in the search space, we evaluate the completeness $\phi_c$ and disjointness $\phi_d$.  Finally we keep the one case with the maximum completeness subject to the constraint that its disjointness is less than $\eta$, as defined in Eqs.~(\ref{eq:completeness}) and~(\ref{eq:disjointness}). The completeness $\phi_c$ for this case is then taken as the matching score between $U$ and $V$ and we denote this score as $\phi(U, V)$. The higher this matching score, the better the partial-to-global matching between $U$ and $V$. Since we consider the cases with $K=1$ and $K=2$ components into one unified optimization process, this algorithm can automatically identify whether $U$ partially contains one or multiple copies of the same design. The process of the composite pattern matching is illustrated in Fig.~\ref{fig:candidate combination} and this algorithm is summarized in Algorithm~\ref{algorithm:flow}.

\begin{algorithm}
\caption{Algorithm for composite sherd-to-design matching.} \label{algorithm:flow}
\begin{algorithmic}[1]
\STATE Input: A sherd image and the design image database 
\FOR {all design images in design image database}
\STATE Extract the curve patterns $U$ from the sherd image and $V$ from a design image 
\FOR {all translation $\bf t$ of $U$ on $V$}
\FOR {all $\theta$ in $[0^{\circ}, 360^{\circ})$ } 
    \STATE  Calculate component $U({\bf T})$ with Chamfer distance $d_{CM}(U_{\bf T}({\bf T}), V)$ 
\ENDFOR
	\STATE Construct a candidate component $U({\bf t})$ by Eq.~(\ref{eq:candidate component construction})
\ENDFOR
\STATE Reduce the candidate components by taking the local minimums at the new distance map $\tilde{d}$
\STATE Find the optimal component $U_i$ or combined components $\{U_i, U_j\}$ from the constrained set defined by Eq.~(\ref{eq:disjointness}), with the maximum completeness $\phi_c$ defined in Eq.~(\ref{eq:completeness}) 
\STATE Store completeness $\phi_c$ as the matching score
\ENDFOR
\STATE Sort the matching scores for all designs and find the best matched designs
\end{algorithmic}
\end{algorithm}

\section{Experiment Results}\label{sec:experiments}

\begin{figure}[htbp]
\begin{center}
   \includegraphics[width=1\linewidth]{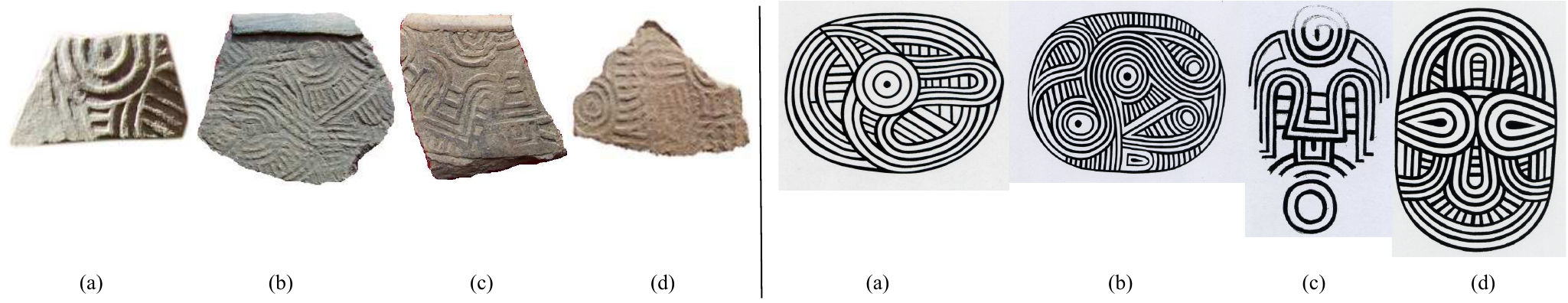}
\end{center}
\caption{(a–d)Sample sherds and designs in our dataset that are used for performance evaluation. Original designs reproduced with permission, courtesy of Frankie Snow, South Georgia State College.}
\label{fig:sample sherds and designs}
\end{figure}

To test the proposed method, we assembled an image dataset of 100 sherds from archaeological sites associated with the Swift Creek paddle stamped tradition of southeastern North America~\cite{stephenson2002aspects,Riggs2003}. These 100 sherds have curved patterns representing 20 unique paddle designs, which have been nearly or fully reconstructed by Frankie Snow from this sherd evidence and others~\cite{snow1975swift}. The curve pattern on each sherd comes from a single design while the same design may be present on multiple sherds. Samples of these sherds and designs are illustrated in Fig.~\ref{fig:sample sherds and designs}. About 80\% of the sherds clearly show composite patterns, e.g., sherds shown in Fig.~\ref{fig:sample sherds and designs}(b), (c) and (d) contain the multiple copies of the same design with spatial overlaps. 

\begin{figure}[htbp]
\begin{center}
 \includegraphics[width=0.9\linewidth]{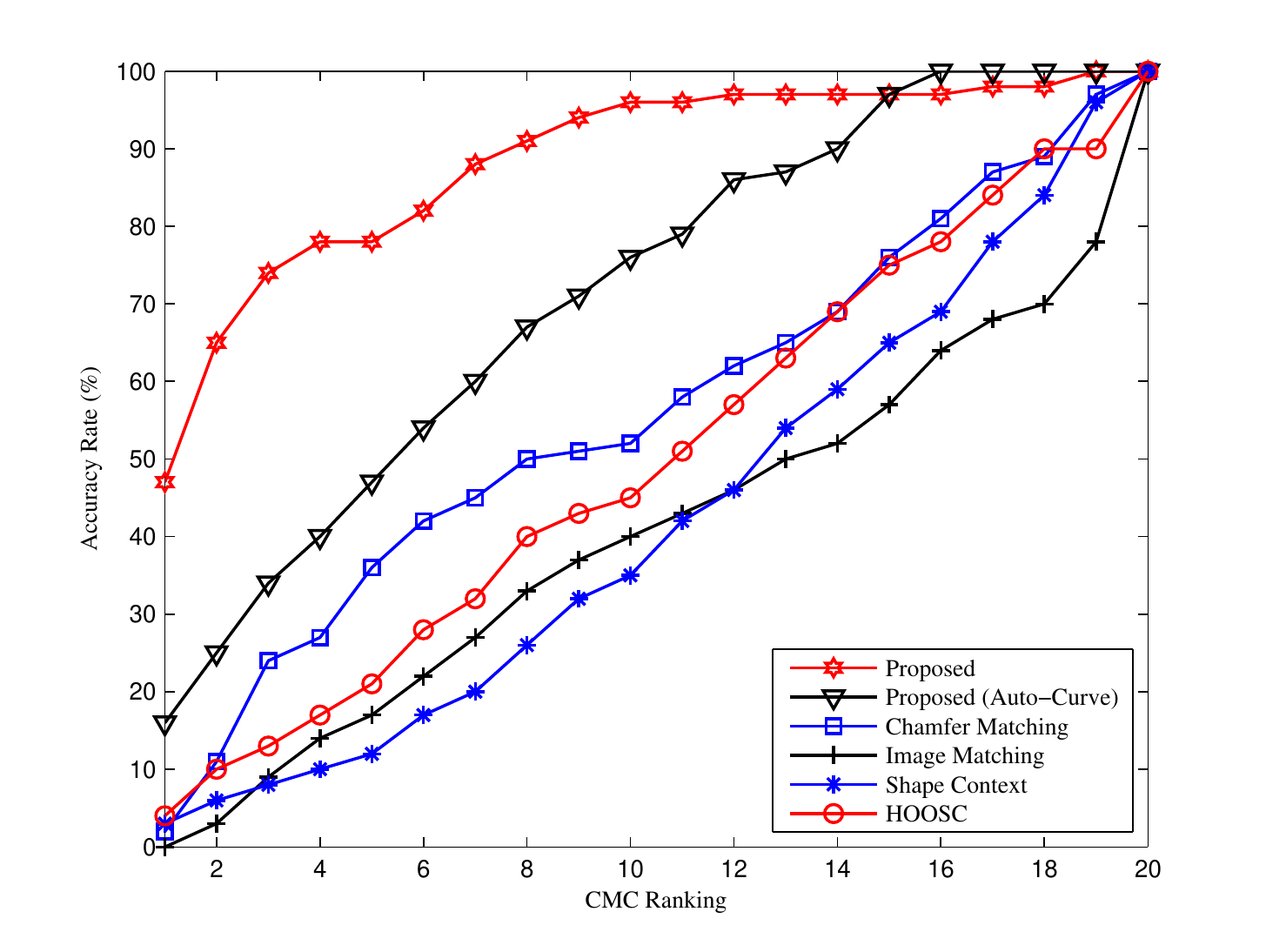}    
\end{center}
   \caption{CMC curves of the proposed method and the four comparison methods. ``Proposed (Auto-Curve)" indicates the performance of the proposed method on the sherd curve patterns extracted without manual refinement.}
\label{fig:composite sherd experiment result}
\end{figure}

In our experiments, we use the Cumulative Matching Characteristics (CMC) ranking metric to evaluate the matching performance. To identify the underlying design of a sherd pattern $U$, we match it against all 20 designs. We then sort these 20 designs in terms of the matching scores and pick the top $L$ designs with the highest scores. If the ground-truth design of a sherd is among the identified top $L$ designs, we treat it as a correct design identification under rank $L$.  We repeat this identification for all $100$ sherds and calculate the accuracy, i.e., the percentage of the correctly identified sherds, under each rank $L$, $L=1, 2,\cdots, 20$. This way, we can obtain a CMC curve in terms of rank $L$ as shown in Fig.~\ref{fig:composite sherd experiment result} to evaluate the performance of each matching algorithm. The higher value in this curve, the better the matching performance.  For parameter settings in the proposed method, we set $\alpha = 3$ to decide whether an edge pixel in the transformed sherd pattern has been matched to an edge pixel in the design (see Section~\ref{sec:composite pattern matching}). We set $\eta=0.1$ for the disjointness score to allow the possible sharing of the some edge pixels between different components in the sherd pattern (see Section~\ref{sec:composite pattern matching}).

To justify the effectiveness of the proposed method, we selected four traditional matching algorithms for performance comparison in the experiments. 1) \emph{(Baseline) Chamfer Matching} without considering composite patterns. It takes the same curve patterns $U$ and $V$ extracted from sherds and designs respectively as in the proposed method, calculates the matching distance 
$d_{CM}(U_{{\bf T}^*},V)$ and uses it for computing the CMC ranking.
2) \emph{Image matching}, in which an one-pixel-wide curve pattern image extracted from a sherd is translated and rotated for a best match to each design curve pattern image in terms of pixel intensity. Denote $I^U$ and $I^V$ as a sherd curve pattern image and a design curve pattern image, respectively, and let $\bf T$ be the transform of $I^U$ over $I^V$, then the matching score $S$ is defined as
\begin{equation}
S = \max_{\bf T}\frac{\sum_{x,y} I^U_{\bf T}(x,y)\cdot I^V(x,y)}{\sqrt{\sum_{x,y} (I^U_{\bf T}(x,y))^2\cdot\sum_{x,y} (I^V(x,y))^2}},
\label{eq:image matching cost}
\end{equation}
where the transform $\bf T$ considers all the possible translation and rotation as in the proposed method. In the experiment, we directly use an OpenCV implementation of this algorithm. 
3) \emph{Shape Context}. It uses the same curve patterns $U$ and $V$ as in the proposed method.  To deal with partial matching,  we use a sliding window technique to match a sherd curve pattern to each window-cropped design curve pattern and then choose the one with the lowest matching distance. The sliding-window size is the same as the sherd image. The Shape Context algorithm implementation directly comes from the OpenCV package. 
4) \emph{Histogram of Orientation Shape Context (HOOSC) }~\cite{roman2011analyzing}. Its setup is the same as Shape Context, but it  incorporates the orientation measure into the log-polar histograms. It is based on Roman-Rangel's~\cite{roman2011analyzing} paper and implemented by HG Zhao\footnote[1]{https://github.com/CyberZHG/Sketch-Based/tree/master/HOOSC} using MATLAB. %

Figure~\ref{fig:composite sherd experiment result} shows the CMC curves of the proposed method and the four comparison methods. Clearly, all four comparison methods show very poor performance by having CMC curves along the diagonal line. The major reason for their poor performance is that they do not consider and cannot well handle the composite patterns present on the sherds. By explicitly considering the possible composite patterns, the proposed method achieves much better CMC performance. In Figure~\ref{fig:composite sherd experiment result}, we also include the CMC curve of ``Proposed (Auto-Curve)", which is from the proposed method on the sherd curve patterns extracted without manual refinement. We can see that the manual refinement is still necessary to extract high-quality curve patterns from the current sherd images.

\begin{figure}[htbp]
\begin{center}
  \includegraphics[width=1\linewidth]{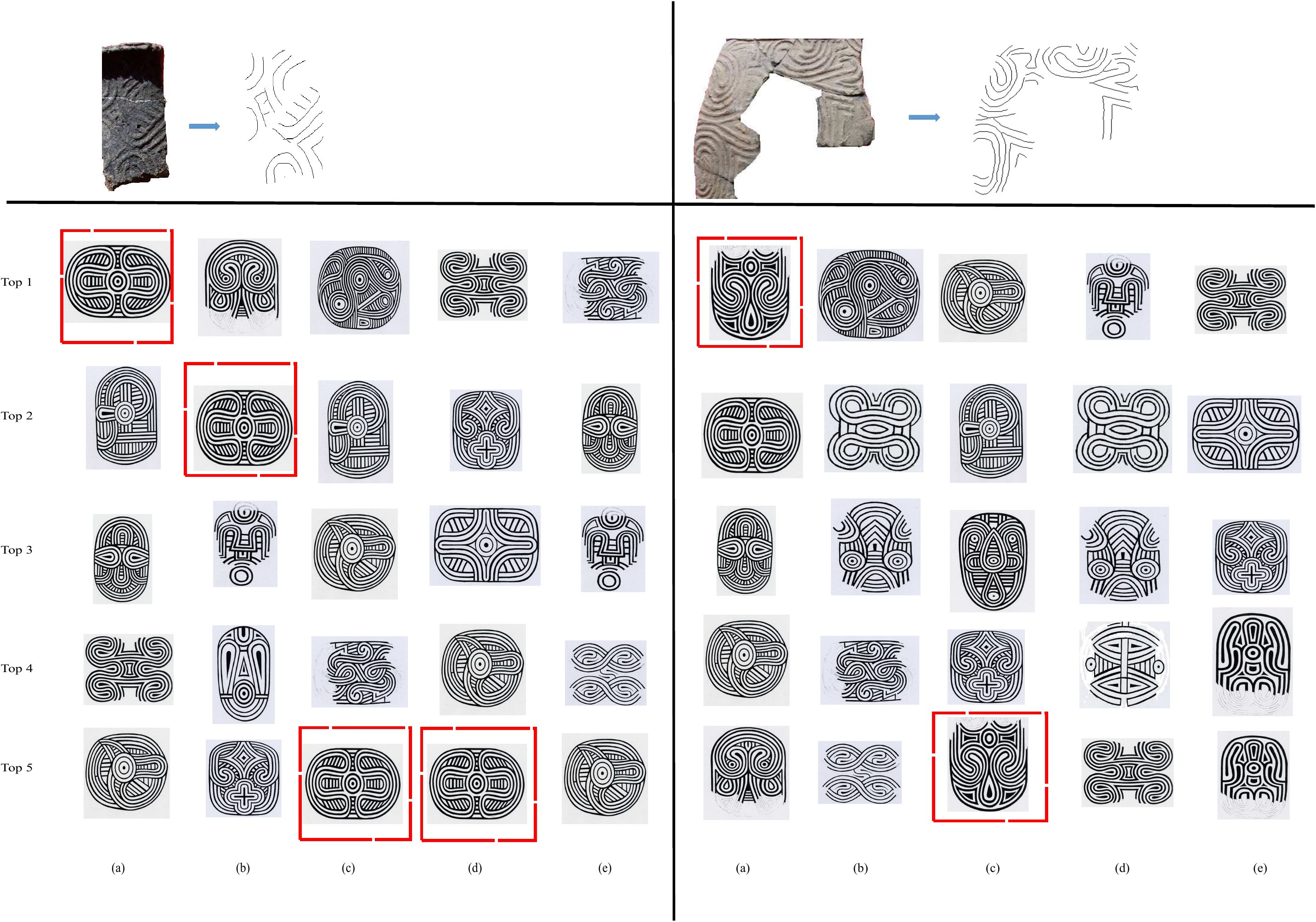}
\end{center}
   \caption{The design identification result for two sample sherds. The top matched designs identified by the proposed method shown in column (a), while the top matched designs identified by Chamfer Matching, Image Matching, Shape Context and HOOSC are shown in columns (b), (c), (d) and (e) respectively. Red boxes indicate the correct designs. Original designs reproduced with permission, courtesy of Frankie Snow, South Georgia State College. }
\label{fig:composite sherd matching examples}
\end{figure}

Figure.~\ref{fig:composite sherd matching examples} shows the sample results of the proposed method and the four comparison methods when matching two sherds to the designs. We can see that, in these two examples, the proposed method can identify the correct designs (in red box) under CMC rank 1, while the four comparison method can only identify the correct designs under much higher CMC ranks.

\begin{figure}[htbp]
\begin{center}
   \includegraphics[width=0.8\linewidth]{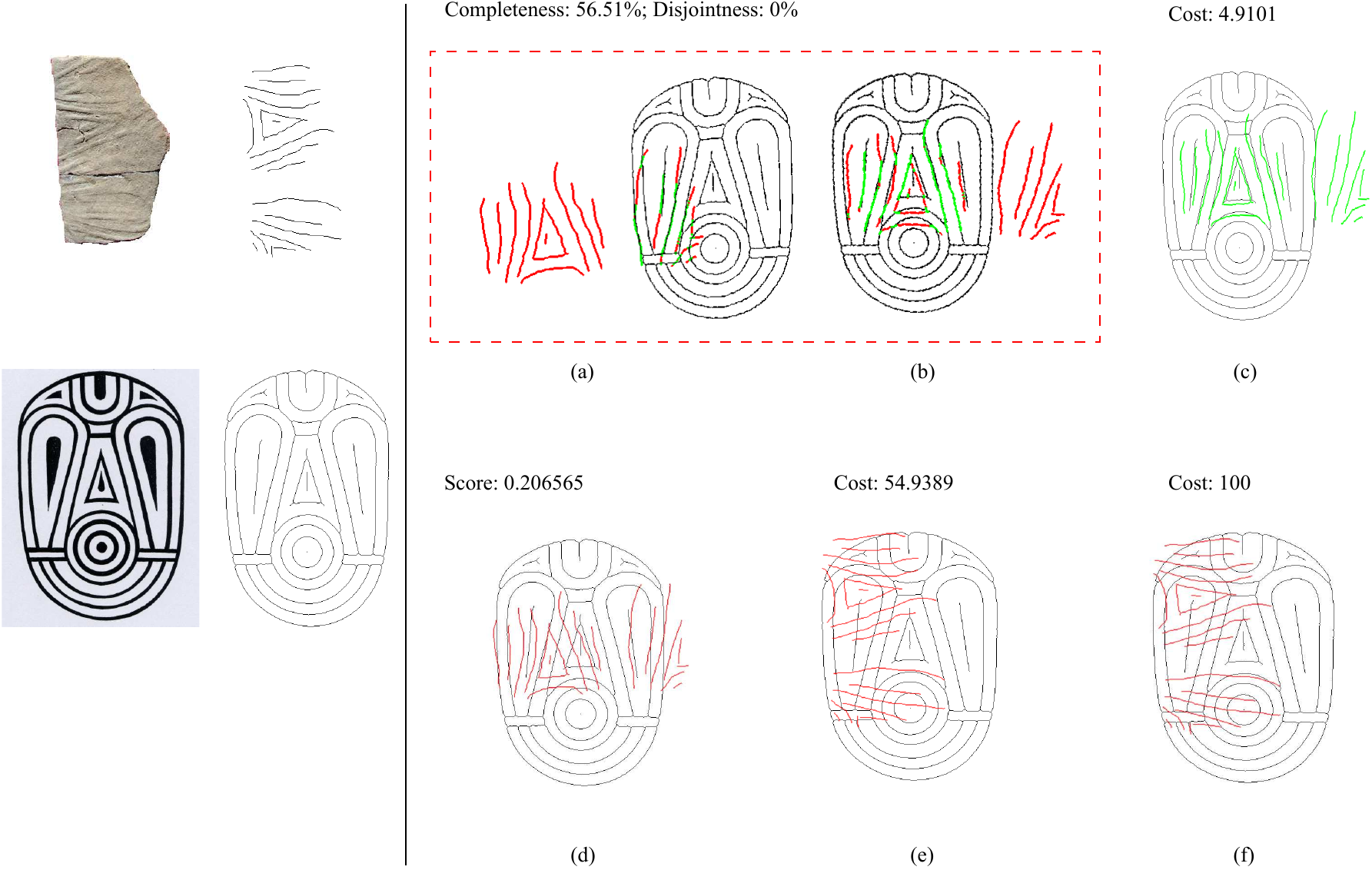} 
\end{center}
   \caption{Matching results of a sample sherd. (a-b) Two matched components on the design using the proposed method. (c) Result from Chamfer Matching. (d) Result from Image Matching. (e) Result from Shape Context. (f) Result from HOOSC. Original design reproduced with permission, courtesy of Frankie Snow, South Georgia State College.}
\label{fig:sherd pattern detection sample 1}
\end{figure}

\begin{figure}[htbp]
\begin{center}
   \includegraphics[width=0.8\linewidth]{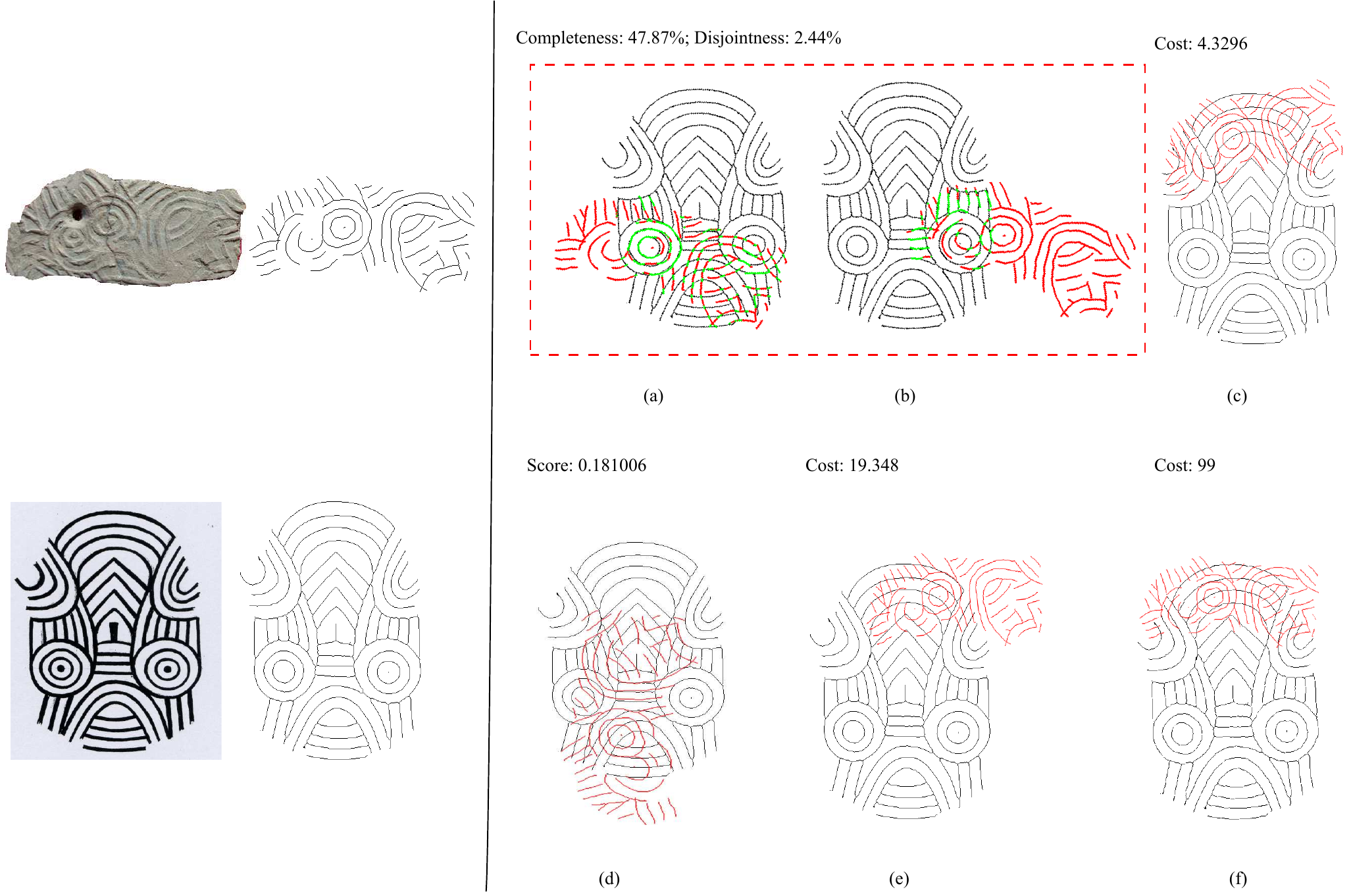}    
\end{center}
   \caption{Matching results of another (second) sample sherd. (a-b) Two matched components on the design using the proposed method. (c) Result from Chamfer Matching. (d) Result from Image Matching. (e) Result from Shape Context. (f) Result from HOOSC. Original design reproduced with permission, courtesy of Frankie Snow, South Georgia State College.}
\label{fig:sherd pattern detection sample 2}
\end{figure}

\begin{figure}[htbp]
\begin{center}
   \includegraphics[width=0.8\linewidth]{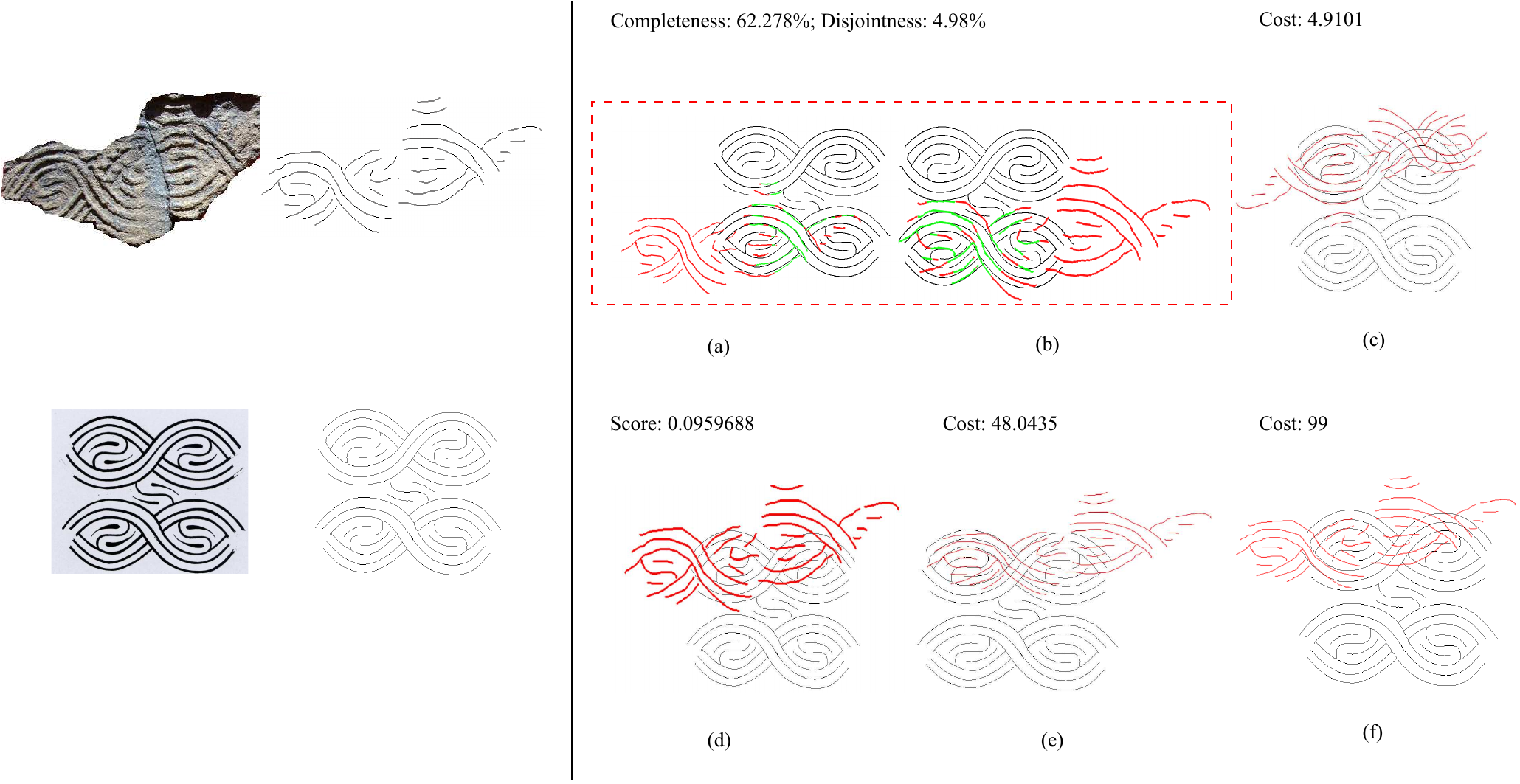}    
\end{center}
   \caption{Matching results of another (third) sample sherd. (a-b) Two matched components on the design using the proposed method. (c) Result from Chamfer Matching. (d) Result from Image Matching. (e) Result from Shape Context. (f) Result from HOOSC. Original design reproduced with permission, courtesy of Frankie Snow, South Georgia State College.}
\label{fig:sherd pattern detection sample 3}
\end{figure}

\begin{figure}[htbp]
\begin{center}
   \includegraphics[width=0.8\linewidth]{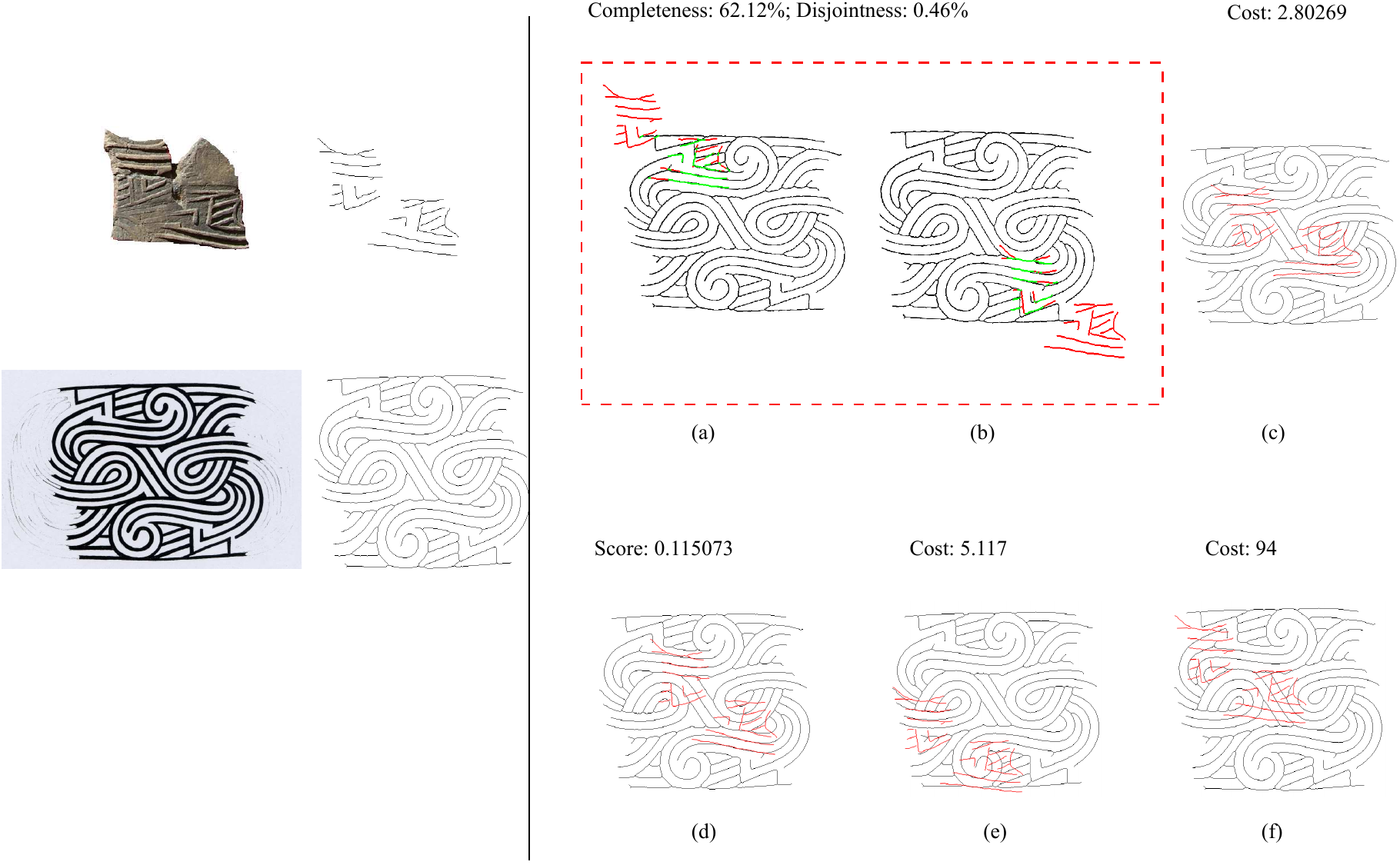}    
\end{center}
   \caption{Matching results of another (fourth) sample sherd. (a-b) Two matched components on the design using the proposed method. (c) Result from Chamfer Matching. (d) Result from Image Matching. (e) Result from Shape Context. (f) Result from HOOSC. Original design reproduced with permission, courtesy of Frankie Snow, South Georgia State College.}
\label{fig:sherd pattern detection sample 4}
\end{figure}

Figures~\ref{fig:sherd pattern detection sample 1}, ~\ref{fig:sherd pattern detection sample 2}, ~\ref{fig:sherd pattern detection sample 3} and ~\ref{fig:sherd pattern detection sample 4} show the matching results of four sherd samples, respectively. On the left column of these figures are the sample sherds and their best matched designs. We can see that composite patterns are present on all four sherds.  
On the right side of these figures, we show the matching results of each sherd over its best matched design, using the proposed method and the four comparison methods. Specifically, for each sherd, (a) and (b) show the identified two components (in green) of the sherd pattern and their matched locations/orientations on the design, respectively. (c), (d), (e) and (f) show the best matched locations/orientations of the sherd pattern on the design using Chamfer Matching, Image Matching, Shape Context and HOOSC respectively. The values above the results of the comparison methods are their respective matching costs or scores. 
For Chamfer Matching, this value is the Chamfer matching distance defined in Eq.~(\ref{eq:best_distance}). For Image Matching, it is the matching score defined in Eq.~(\ref{eq:image matching cost}). For Shape Context and HOOSC, these values are their respective matching distances. 
Note that, in the four comparison methods, the sherd pattern is not decomposed into multiple components and they are matched to the design as a whole. Archaeologists who specialize in the study of these designs confirmed that the matching results of these four sherds are correct when using the proposed method. For the comparison methods, the only correct matching is produced by Chamfer Matching on the first sample sherd, as shown in Fig.~\ref{fig:sherd pattern detection sample 1} (c).

\begin{figure}[htbp]
\begin{center}

   \includegraphics[width=1\linewidth]{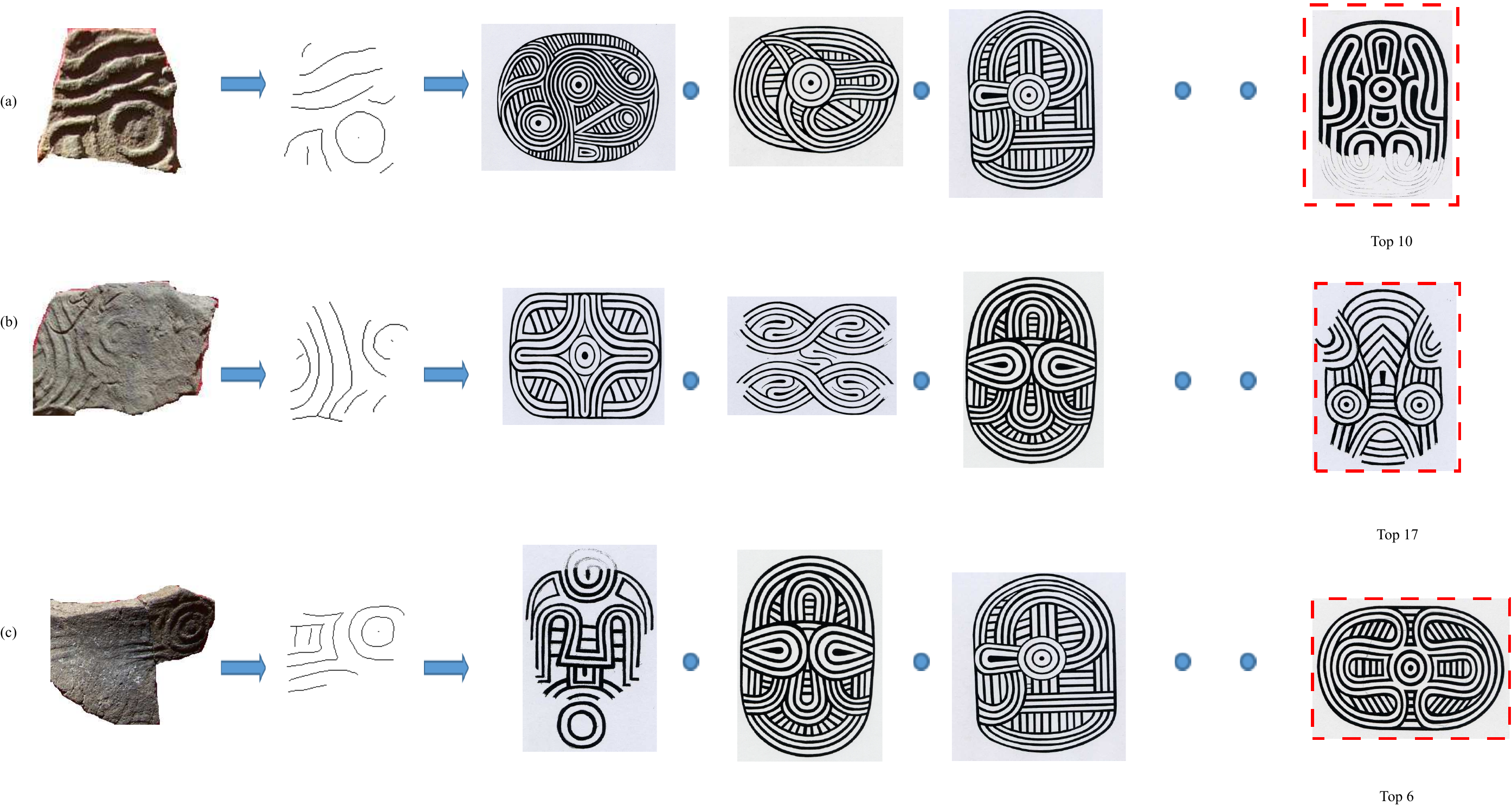}
\end{center}
   \caption{Three failure cases where the top five matched designs do not include the ground-truth design. From left to right are the sherd image, curve pattern extracted from the sherd and the top matched designs returned by the proposed method, respectively. The last column shows the ground-truth design for each sherd and its CMC rank (below each ground-truth design). Original designs reproduced with permission, courtesy of Frankie Snow, South Georgia State College.}
\label{fig:failure cases}
\end{figure}

We also inspected the experiment results to find the failure cases when using the proposed method and the cause of failure cases. Specifically, we examined the sherds whose ground-truth designs are not among the top five matchings, i.e., incorrect matching under CMC rank 5. Examples of these failure cases are shown in Fig.~\ref{fig:failure cases}.

We found that most of these failure cases are caused by the local pattern similarity of the designs. Most of the southeastern North America paddle designs are usually combinations of simple curve patterns, such as the concentric circles as shown in Fig.~\ref{fig:failure cases}(a). When the composition of a sherd pattern is dominated by such simple curve patterns, it can be easily confused to many designs other than its ground truth design. Furthermore, the curve pattern extracted from a real sherd usually contains noise, missing segments, and inaccuracies because of variable surface smoothing during vessel manufacture, incomplete application of the planar paddle to the curved pottery surface, and surface erosion from post-depositional weathering as shown in Fig.~\ref{fig:failure cases}(b). Another possible issue is the deformation when using a perspective camera to take the sherd image from different view angles as shown in Fig.~\ref{fig:failure cases}(c), although we try our best to take the picture perpendicular to the center of sherd surface. Such image deformation may be reduced by customizing the camera setup to ensure its perpendicularity to the center of sherd surface.

\section{Conclusion}\label{sec:conclusion}

In this paper, we developed a new partial-to-global curve-pattern matching algorithm to identify the designs of the carved wooden paddles from unearthed pottery sherds. Different from previous partial matching problems, the curve pattern on each sherd may be a composite one resulting from multiple, partially overlapped copies of the same design. To address this problem, we extended the classical Chamfer matching to identify candidate components of the sherd pattern and then leveraged two metrics of completeness and disjointness to find the optimal sherd-pattern decomposition.  In the experiment, we tested a collection of 100 sherds against 20 known southeastern North America paddle designs. The results show that the CMC performance of the proposed method is substantially better than several traditional image and curve-pattern matching algorithms.

\section*{Acknowledgement}\label{sec:acknowledgement}
This research is supported by National Center for Preservation Technology and Training Grants Program (P16AP00373) and University of South Carolina Social Sciences Grant Program. We would like to show our gratitude to Professor Frankie Snow at South Georgia State College for sharing his pearls of wisdom and design images with us during the course of this research. We also thank our colleague Scot Keith for encouraging the pursuit of this research.

\bibliography{report}   
\bibliographystyle{spiejour}   


\vspace{2ex}\noindent\textbf{Jun Zhou} is a Ph.D. Candidate in Department of Computer Science and Engineering at University of South Carolina. She received her B.S. degree in Computer Science at Nankai university in China and M.S. degree in Computer Science at Loyola University at Chicago in 1996 and 2000 respectively. Her current interests include document image processing and cultural heritage object image processing. 

\vspace{2ex}\noindent\textbf{Haozhou Yu} is a Ph.D. student in Department of Computer Science and Engineering at University of South Carolina. He received his B.S. degree from the Department of Bio-medical Engineering at Beijing Jiaotong University. His interests include human tracking, cultural heritage object image processing. 

\vspace{2ex}\noindent\textbf{Karen Smith} is a southeastern archaeologist with a background in Woodland period and plantation-era research and archaeological curation. She works as the Director of Applied Research Division, South Carolina Institute of Archaeology and Anthropology at University of South Carolina. Her interests include Woodland period and Plantation research and archaeological data analysis. Karen holds a Ph.D. from the University of Missouri (2009).

\vspace{2ex}\noindent\textbf{Colin Wilder} is the Associate Director of Center for Digital Humanities and the director of the Republic of Literature and of the Dirty History Metacrawler at the University of South Carolina. He obtained a B.A. degree in Philosophy from Yale and Ph.D. degree in German history from University of Chicago in 2010. His research focuses on the development of ideas of liberty and equality in German and broader European history in the Early Modern period. 

\vspace{2ex}\noindent\textbf{Hongkai Yu} is a Ph.D. candidate in Department of Computer Science and Engineering at University of South Carolina. He received his M.S. and B.S. degrees from the Department of Automation with a minor in Traffic Control at Changan University, Xi'an, China in 2012 and 2009 respectively. His current research interests include computer vision, machine learning and intelligent transportation system. He is a student member of the IEEE.  

\vspace{2ex}\noindent\textbf{Song Wang} received a Ph.D. degree in electrical and computer engineering from the University of Illinois at Urbana-Champaign in 2002. He is currently a Professor at the Department of Computer Science and Engineering, University of South Carolina. His current research interests include computer vision, image processing, and machine learning. He serves as an Associate Editor of Pattern Recognition Letters. He is a senior member of IEEE and a member of the IEEE Computer Society.

\end{spacing}
\newpage
\section*{List of Figure Captions}

Fig.~\ref{fig:sample designs}: Five paddle designs reconstructed by Frankie Snow. Original design reproduced with permission, courtesy of Frankie Snow, South Georgia State College.

Fig.~\ref{fig:sample pottery sherds}: Sample pottery sherds (top) and their underlying wooden paddle designs (bottom). Two pottery sherds in (b) contain a composite pattern, resulting from the multiple applications of the carved paddle with partial spatial overlaps. Original designs reproduced with permission, courtesy of Frankie Snow, South Georgia State College.

Fig.~\ref{fig:non-composite sherd-to-design matching procedure}: An illustration of the procedure of identifying the underlying design for a sherd: first extracting the curve pattern on the sherd, which is then matched to each design in a database of known designs for identifying the best matched design. Original design reproduced with permission, courtesy of Frankie Snow, South Georgia State College.

Fig.~\ref{fig:curve from a design}: An illustration of curve extraction from a sherd and a design. (a) Curve extraction from a sherd. (b) Curve extraction from a design. Original design reproduced with permission, courtesy of Frankie Snow, South Georgia State College.

Fig.~\ref{fig:curve extraction}: An illustration of curve pattern extraction from a sherd. 1) Converting color image to gray-scale image. 2) Image enhancement. 3) Ridge detection. 4) Thresholding for binary ridge image. 5) Noise removal. 6) Thinning. 7) Short branch removal. 8) Manual refinement (if needed).

Fig.~\ref{fig:failure cases on curve extraction}: Sample curve extraction results from automatic image processing, i.e., Steps 1) through 7). They need substantial manual refinement.

Fig.~\ref{fig:distance map}: An illustration of the distance map and Chamfer matching. (a) Curve pattern on a sherd. (b) Distance map of a design - brighter pixels indicate higher values in the distance map. (c) Chamfer matching result (in red). From original design by Frankie Snow, South Georgia State College.

Fig.~\ref{fig:composite pattern curves matching problem}: An Illustration of a composite pattern, which consists of two components. (a) A sherd with a composite pattern. (b) The extracted composite pattern. (c) The underlying design. (d) Two components (red and green) of the composite pattern matched to different parts of the design, with blue curve fragments shared by two components. From original design by Frankie Snow, South Georgia State College.

Fig.~\ref{fig:candidate combination}: The process of combining candidate components for matching to a design ($K=2$). The optimal result is indicated in the red box. (a) Matching a sherd pattern (top) to a design pattern (bottom). (b) Candidate components. (c) Combining candidate components (completeness scores $\phi_c$ shown in red and disjointness scores $\phi_d$ shown in black). From original design by Frankie Snow, South Georgia State College.

Fig.~\ref{fig:sample sherds and designs}: Sample sherds and designs in our dataset that are used for performance evaluation. Original designs reproduced with permission, courtesy of Frankie Snow, South Georgia State College.

Fig.~\ref{fig:composite sherd experiment result}: CMC curves of the proposed method and the four comparison methods. ``Proposed (Auto-Curve)" indicates the performance of the proposed method on the sherd curve patterns extracted without manual refinement.

Fig.~\ref{fig:composite sherd matching examples}: The design identification result for two sample sherds. The top matched designs identified by the proposed method shown in column (a), while the top matched designs identified by Chamfer Matching, Image Matching, Shape Context and HOOSC are shown in columns (b), (c), (d) and (e) respectively. Red boxes indicate the correct designs. Original designs reproduced with permission, courtesy of Frankie Snow, South Georgia State College.

Fig.~\ref{fig:sherd pattern detection sample 1}: Matching results of a sample sherd. (a-b) Two matched components on the design using the proposed method. (c) Result from Chamfer Matching. (d) Result from Image Matching. (e) Result from Shape Context. (f) Result from HOOSC. Original design reproduced with permission, courtesy of Frankie Snow, South Georgia State College.

Fig.~\ref{fig:sherd pattern detection sample 2}: Matching results of another (second) sample sherd. (a-b) Two matched components on the design using the proposed method. (c) Result from Chamfer Matching. (d) Result from Image Matching. (e) Result from Shape Context. (f) Result from HOOSC. Original design reproduced with permission, courtesy of Frankie Snow, South Georgia State College.

Fig.~\ref{fig:sherd pattern detection sample 3}: Matching results of another (third) sample sherd. (a-b) Two matched components on the design using the proposed method. (c) Result from Chamfer Matching. (d) Result from Image Matching. (e) Result from Shape Context. (f) Result from HOOSC. Original design reproduced with permission, courtesy of Frankie Snow, South Georgia State College.

Fig.~\ref{fig:sherd pattern detection sample 4}: Matching results of another (fourth) sample sherd. (a-b) Two matched components on the design using the proposed method. (c) Result from Chamfer Matching. (d) Result from Image Matching. (e) Result from Shape Context. (f) Result from HOOSC. Original design reproduced with permission, courtesy of Frankie Snow, South Georgia State College.

Fig.~\ref{fig:failure cases}: Three failure cases where the top five matched designs do not include the ground-truth design. From left to right are the sherd image, curve pattern extracted from the sherd and the top matched designs returned by the proposed method, respectively. The last column shows the ground-truth design for each sherd and its CMC rank (below each ground-truth design). Original designs reproduced with permission, courtesy of Frankie Snow, South Georgia State College.

\end{document}